\documentclass{article}

\usepackage{arxiv}
\usepackage[utf8]{inputenc} 
\usepackage[T1]{fontenc}    
\usepackage{hyperref}       
\usepackage{url}            
\usepackage{booktabs}       
\usepackage{amsfonts}       
\usepackage{nicefrac}       
\usepackage{microtype}      
\usepackage{xcolor}
\usepackage{xfrac}
\usepackage{dsfont}
\usepackage{xspace} 
\usepackage{doi}
\usepackage{empheq}
\usepackage[flushleft]{threeparttable}
\usepackage{caption}

\usepackage{mathtools} 
\usepackage{longtable} 
\usepackage{array} 
\usepackage{pifont}
\usepackage{tikz} 
\usepackage{amssymb}
\usepackage{amsthm}
\usepackage{breqn}
\usepackage{bm}
\usepackage{graphicx}
\usepackage{subcaption}
\usepackage[ruled,linesnumbered]{algorithm2e}
\DontPrintSemicolon
\usepackage{cleveref}
\crefname{assumption}{assumption}{Assumptions}
\usepackage{wrapfig}
\usepackage{placeins}
\usepackage[toc,page,header]{appendix}
\usepackage{minitoc}
\usepackage{cases}
\usepackage{siunitx}
\usepackage{diagbox}
\usepackage{multirow}
\usepackage{makecell}
\usepackage{ragged2e}
\usepackage{tcolorbox}
\usepackage[super]{nth}
\usepackage{float}

\newtheorem{definition}{Definition}

\DeclareMathOperator*{\argmax}{arg\,max}

\DeclareMathOperator*{\arginf}{arg\,inf}

\newcommand{\bigO}{\mathcal{O}}

\newcommand{\R}{\mathbb{R}}

\newcommand{\I}{\mathds{1}}
\newcommand{\E}{\mathbb{E}}
\newcommand{\e}{\mathcal{E}}
\newcommand{\p}{\mathbb{P}}

\newcommand{\tf}{\widetilde{f}}

\newcommand{\sis}{{s\in\s}}
\newcommand{\tM}{\widetilde{M}}
\newcommand{\tMn}{\widetilde{M}_n}
\newcommand{\hM}{\widehat{M}}

\newcommand{\A}{\mathcal{A}}
\newcommand{\D}{\mathcal{D}}
\newcommand{\s}{\mathcal{S}}

\newcommand{\at}{a_{t}}
\newcommand{\ut}{u_{t}}
\newcommand{\pt}{p_{t}}
\newcommand{\pit}{\pi_{t}}
\newcommand{\pitn}{\pi_{t+1}}

\newcommand{\pitsmr}{\pit(\st,\mut,\deltat)}

\newcommand{\st}{s_{t}}
\newcommand{\snt}{s_{n,t}}
\newcommand{\stn}{s_{t{}+{}1}}

\newcommand{\mut}{\mu_{t}}

\newcommand{\muant}{\mua_{n,t}}
\newcommand{\bnt}{b_{n,t}}

\newcommand{\omuta}{\overline{\mu}_t^A}

\newcommand{\mutn}{\mu_{t{}+{}1}}

\newcommand{\Umpf}{U(\mut, \etat, \pit, f)}

\newcommand{\hw}{h^w}

\newcommand{\Phit}{\Phi_{t}}
\newcommand{\ds}{ds}
\newcommand{\djs}{\mathrm{D_{JS}}}

\newcommand{\epsilont}{\epsilon_{t}}
\newcommand{\epsilontr}{\epsilont^R}
\newcommand{\epsilontm}{\epsilont^M}
\newcommand{\epsilontc}{\epsilont^C}

\newcommand{\deltat}{\delta_{t}}
\newcommand{\deltant}{\delta_{n,t}}
\newcommand{\odelta}{\overline{\delta}}
\newcommand{\odeltat}{\odelta_t}
\newcommand{\muta}{\mut^A}
\newcommand{\mua}{\mu^A}

\newcommand{\mutr}{\mut^R}

\newcommand{\mutm}{\mut^M}

\newcommand{\mutc}{\mut^C}

\newcommand{\mt}{m_{t}}

\newcommand{\pps}{\pp(\s)}
\newcommand{\bbs}{\bb(\s)}

\newcommand{\mms}{\mm(\s)}

\newcommand{\eq}{{}={}}
\newcommand{\pp}{\mathcal{P}}

\newcommand{\mm}{\mathcal{M}}
\newcommand{\bb}{\mathcal{B}}
\newcommand{\hf}{\widehat{f}}

\newcommand{\bpi}{\boldsymbol{\pi}}
\newcommand{\bpin}{\bpi_n}
\newcommand{\bpis}{\bpi^*}
\newcommand{\bpisot}{\bpis_{OT}}
\newcommand{\bpins}{\bpin^*}

\newcommand{\Pit}{\Pi_t}
\newcommand{\bPi}{\boldsymbol{\Pi}}
\newcommand{\oR}{\overline{r}}

\newcommand{\etat}{\eta_{t}}

\renewcommand{\epsilon}{\ensuremath\varepsilon}

\newcommand{\mfcp}[1]{MFC($#1$)}
\newcommand{\mfrlp}[1]{MFRL($#1$)}

\newcommand{\done}[1]{\textcolor{blue}{DONE}}

\newcommand{\red}[1]{%
  \ifshowchanges
    \textcolor{red}{#1}%
  \else
    #1%
  \fi
}

\definecolor{darkgreen}{rgb}{0.0, 0.5, 0.0}
\newcommand{\darkgreen}[1]{%
  \ifshowchanges
    \textcolor{darkgreen}{#1}%
  \else
    #1%
  \fi
}

\newcommand{\noi}{\noindent}
\newcommand{\cmfmdp}{C-MF-MDP\xspace}
\newcommand{\mdp}{\mathcal{MDP}}

\newcommand{\cmark}{\darkgreen{\ding{51}}} 
\newcommand{\xmark}{\red{\ding{55}}} 
\usepackage[
    backend=biber,%
    style=nature, 
    natbib=true,%
    clearlang=true,%
]{biblatex}

\AtEveryBibitem{%
  \clearfield{day}%
  \clearfield{month}%
  \clearfield{endday}%
  \clearfield{endmonth}%
}

\everypar{\looseness=-1}

\title{
Scalable Ride-Sourcing Vehicle Rebalancing with Service Accessibility Guarantee: A Constrained Mean-Field Reinforcement Learning Approach
}

\date{} 					

\author{
	Matej Jusup \\
	ETH Z\"{u}rich, Switzerland \\
	\texttt{mjusup@ethz.ch} \\
    \And
	Kenan Zhang\\
	EPFL Lausanne, Switzerland \\
	\texttt{kenan.zhang@epfl.ch} \\
    \And
	Zhiyuan Hu\\
	EPFL Lausanne, Switzerland \\
	\texttt{zhiyuan.hu@epfl.ch}
    \\
	\And
	Barna P\'{a}sztor \\
	ETH Z\"{u}rich, Switzerland \\
	\texttt{bpasztor@ethz.ch} \\
    \And
	Andreas Krause\\
	ETH Z\"{u}rich, Switzerland \\
	\texttt{krausea@ethz.ch} \\
    \And
	Francesco Corman\\
	ETH Z\"{u}rich, Switzerland \\
	\texttt{corman@ethz.ch} \\
}

\renewcommand{\headeright}{}
\renewcommand{\undertitle}{}
\renewcommand{\shorttitle}{Ride-Sourcing Vehicle Rebalancing with Service Accessibility Guarantee: A Mean-Field RL Approach}

\newif\ifshowchanges
\showchangestrue   

\begin{document}
\maketitle
\begin{abstract}
The rapid expansion of ride-sourcing services such as Uber, Lyft, and Didi Chuxing has fundamentally reshaped urban transportation by offering flexible, on-demand mobility via mobile applications. Despite their convenience, these platforms confront significant operational challenges, particularly vehicle rebalancing---the strategic repositioning of a large group of vehicles to address spatiotemporal mismatches in supply and demand. Inadequate rebalancing not only results in prolonged rider waiting times and inefficient vehicle utilization but also leads to fairness issues, such as the inequitable distribution of service quality and disparities in driver income. 
To tackle these complexities, we introduce continuous-state mean-field control (MFC) and mean-field reinforcement learning (MFRL) models that employ continuous vehicle repositioning actions. MFC and MFRL offer scalable solutions by modeling each vehicle's behavior through interaction with the vehicle distribution, rather than with individual vehicles.
This mitigates the curse of dimensionality with respect to the number of agents, enabling coordination across large fleets with significantly reduced computational complexity and eliminating the need to retrain the model when fleet size changes.
To ensure equitable service access across geographic regions, we integrate an accessibility constraint into both models and derive rebalancing policies that strike a balance between high fulfillment of passenger demand and fair coverage of vehicle supply.
Extensive empirical evaluation using real-world data-driven simulation of Shenzhen demonstrates the real-time efficiency and robustness of our approach. 
Remarkably, it scales to tens of thousands of vehicles, with training times comparable to the decision time of a linear programming rebalancing.
Besides, policies generated by our approach effectively explore the efficiency-equity Pareto front, outperforming conventional benchmarks across key metrics like fleet utilization, fulfilled requests, and pickup distance, while ensuring equitable service access. \looseness=-1

\end{abstract}


\section{Introduction} \label{sec:introduction}

\begin{figure}[htb!]
\centering
 \includegraphics[width=0.5\columnwidth, trim=11.2cm 7.9cm 10cm 6.4cm, clip]{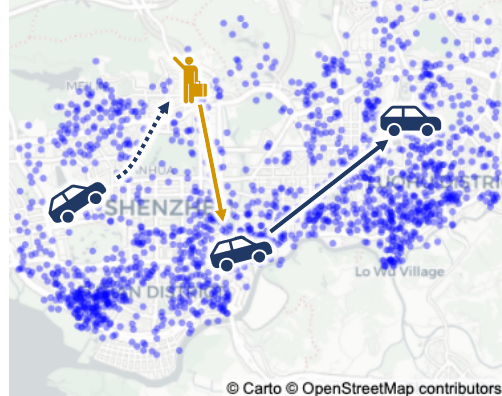}
 \caption{An illustration of vehicle distribution (light-blue scatters), vehicle cruising (dashed blue arrow), serving a rider (yellow arrow), and vehicle repositioning (solid blue arrow). }
 \label{fig:repositioning-trajectory}
\end{figure}

The past decade has witnessed rapid growth in personal mobility services, pioneered by ride-sourcing services (e.g., Uber, Lyft, and Didi Chuxing) that provide individual riders with convenient and efficient mobility solutions. 
Compared to conventional ride-hailing services (e.g., taxis), ride-sourcing is advanced in several aspects. First, it enables matching between riders and vehicles in real-time, which largely reduces the matching friction that has long restricted the operational efficiency of ride-hailing services~\citep{cramer2016disruptive}. 
Besides trip dispatching, other operational strategies (e.g., pricing, routing, incentives) are also optimized dynamically in ride-sourcing services, or even customized for the users~\citep{wang2019ridesourcing}. 
In particular, vacant vehicles' rebalancing has been examined extensively in both the literature and real practice~\citep{zhang2016control,braverman2019empty,jiao2021real}. 
In short, vehicle rebalancing describes a control strategy that adaptively moves vacant vehicles across regions to address the inherent spatiotemporal imbalance between demand and supply in the ride-sourcing market, as illustrated in \Cref{fig:repositioning-trajectory}. It thus benefits both drivers and riders with shorter searching and waiting times, respectively, as well as the ride-sourcing platform with service rate and revenue. \looseness=-1

Vehicle rebalancing is widely known as a challenging problem due to several uncertainties and computational complexity. 
First and foremost, since it takes time for vehicles to travel from one service zone to another, rebalancing decisions must be made in anticipation of future demand. This becomes more critical if vehicles are not allowed to take new requests before finishing their rebalancing trips~\citep{yang2022learning}. 
Secondly, the number of repositioning vehicles (we reserve the term repositioning for a single vehicle and rebalancing for the fleet) is hard to determine due to the complex vehicle-rider matching mechanism. Although equipped with advanced algorithms, matching in ride-sourcing is never frictionless---within a short period, the number of pickups is usually less than the minimum between the number of waiting riders and the number of vacant vehicles. Further, the matching probability, i.e., the probability of picking up a rider within a search period for each driver, is influenced by a number of factors, ranging from the network topology to the trip dispatching algorithm~\citep{zhang2023ride}. 
Last but not least, as vehicle rebalancing is executed in real-time, the solution algorithm must strike a balance between optimality and computational efficiency. 

Recently, deep reinforcement learning (DRL) emerged as a promising solution to vehicle rebalancing~\citep{oda2018movi,mao2020dispatch}. Equipped with deep neural networks, DRL is capable of learning high-dimensional and complicated dynamics between demand and supply in ride-sourcing systems. 
However, the centralized vehicle rebalancing controlled by a global RL agent suffers from scalability issues as the complexity of the problem scales exponentially with the number of vehicles in the network~\citep{lin2018efficient}
On the other hand, multi-agent reinforcement learning (MARL) distributes the rebalancing decisions to individual vehicles, thus largely resolving the curse of dimensionality by reducing the scaling complexity from exponential to linear. 
Yet, MARL remains challenging to scale to a large number of agents and induces a non-stationary environment.
Non-stationarity makes MARL suffer from convergence issues as vehicles learn their individual repositioning strategies concurrently and have to adapt to each other's behavior in the network constantly. In most cases, it is not guaranteed that individually learning vehicles can find jointly stable and optimal strategies~\citep{shou2020reward}.
To stabilize the learning process, coordination mechanisms have been introduced to allow agents to exchange information with others, which again causes the scalability issue due to the increasing complexity of available actions to each vehicle, as they have to devise efficient communication with other vehicles.
This issue motivates the development of vehicle rebalancing approaches based on mean-field reinforcement learning (MFRL), which exploits the mean-field theory from physics to simplify the representation of individual agents~\citep{yang2018mean}. MFRL focuses on interactions between a vehicle and the aggregate behavior of the fleet rather than on individual interactions with every vehicle, which considerably simplifies the optimization space.
MFRL, as well as the general mean-field approach, has also been utilized in recent studies on ride-sourcing beyond the problem of vehicle rebalancing, such as trip dispatching~\citep{li2019efficient}, vehicle routing~\citep{shou2020optimal,zhang2023ride}, and ride pricing~\citep{zhu2021mean}.

Existing vehicle rebalancing approaches mostly focus on optimizing system-wide metrics such as the number of served riders (service rate) or revenue.
As a result, vacant vehicles are usually repositioned to high-demand areas to most effectively improve efficiency, i.e., shorter rider waiting time and driver searching time, and higher matching rate and trip revenue \citep{grahn2020socioeconomic}.
This is in contrast to public transit, which, instead of chasing economic performance, aims to provide universal service even in low-demand areas. 
We argue that an ideal ride-sourcing system should show a high degree of responsiveness towards variations in demand, which cannot be well handled by traditional public transport with fixed routes and schedules, while ensuring a minimum level of service accessibility everywhere, regardless of demand. 
The service accessibility guarantee helps address the equity issue argued by the public in recent years regarding private ride-sourcing companies~\citep{hughes2016transportation,ge2020racial}, though, to the best of our knowledge, it has not been thoroughly investigated and addressed in the literature.
From the modeling perspective, the minimum service accessibility imposes additional constraints on the vehicle rebalancing problem. It thus adds another layer of complexity, particularly for RL-based approaches, as constraints are notoriously difficult to tackle in the RL framework. 
Nevertheless, the recent development of safe RL~\citep{garcia2015comprehensive,gu2022review} also sheds light on how to deal with the constraints in vehicle rebalancing. We contribute to the vehicle rebalancing methodologies by successfully integrating accessibility constraints into the mean-field approaches.

Motivated by the long-standing scalability challenge and the emerging consideration of service accessibility, this paper proposes a constrained mean-field reinforcement learning approach to ensure service accessibility guarantees in the ride-sourcing vehicle rebalancing problem. As an intermediate step, we also develop a mean-field control (MFC) method based on presumed system dynamics. The minimum service accessibility constraint is embedded in both the MFRL and MFC frameworks as a population-level constraint. 
This paper contributes to the literature on vehicle rebalancing with both novel methodologies and practical solutions as follows: 
\begin{itemize}
    \item To the best of our knowledge, this is the first study that has integrated the service accessibility constraint into RL-based vehicle rebalancing under stochastic transitions and non-stationary demand. Since the proposed learning framework is flexible enough to include other constraints, it also has great potential to be applied to a wide range of operational problems in ride-sourcing services.
    \item Unlike most prior studies that control vehicle movements at the zone level, the MFC and MFRL approaches developed in this study reposition vehicles on a continuous space while scaling to tens of thousands of vehicles. The model-based framework also ensures a higher sample efficiency. 
    \item Our simulation experiments on real-world networks demonstrate that MFC and MFRL policies can more efficiently generate rebalancing actions and outperform benchmarks with significant margins. The scalability is rooted in the mean-field approximation, under which no retraining is needed when the fleet size changes. Moreover, the two policies show strong robustness towards demand variations. 
\end{itemize}

The remainder of this paper is organized as follows. In \Cref{sec:review}, we review the previous studies on ride-sourcing vehicle rebalancing and fleet management and provide a brief overview of safe RL, MFC, and MFRL. 
\Cref{sec:model-setup} formally describes the constrained vehicle rebalancing problem, along with the standing assumptions used throughout this paper. 
In \Cref{sec:mfc-and-mfrl}, we first formulate the rebalancing problem in the framework of MFC and then identify the learning tasks in the corresponding MFRL framework. 
\Cref{sec:experiments} describes the simulation environment, benchmark rebalancing policies, key metrics for performance evaluation, and main findings and insights generated from the results. 
Finally, \Cref{sec:conclusion} concludes this paper and provides directions for future research.

\section{Related Work}\label{sec:review}
\subsection{Vehicle Rebalancing and Fleet Management} \label{sec:vehicle-rebalancing-and-fleet-management}
Vehicle rebalancing in ride-sourcing services refers to the movements of a large group of vacant vehicles across regions to balance the demand and supply in the market. It is thus different from individual vehicle repositioning discussed in another group of studies that aim to maximize a single vehicle's performance metrics (e.g., pickup probability and hourly revenue) \citep{yu2019markov,shou2020optimal}. In contrast, vehicle rebalancing has the objective of optimizing system-wide performance (e.g., total trip revenue and average vehicle occupancy rate) and is thus also referred to as fleet management. 
Early studies on vehicle rebalancing and fleet management model the system dynamics as a queuing network and optimize the rebalancing vehicle flows based on the fluid model at the stationary state~\citep{pavone2012robotic,zhang2016control,braverman2019empty}. 
To incorporate uncertainty, stochastic and robust optimization frameworks have been employed~\citep{miao2017data, guo2021robust, jung2019large}. Alternative approaches, including hand-crafted heuristics~\citep{jung2019large} and model predictive control~\citep{tsao2019model}, have also been explored. However, these methods 
typically require explicit modeling of system dynamics and solving optimization problems in real time, which limits their adaptability and scalability in real-world applications.

Recently, a growing line of research has approached the vehicle rebalancing problem using reinforcement learning (RL)~\citep{qin2022reinforcement}. \Cref{tab:literature-review} provides a comprehensive overview of these studies. It is also worth noting that RL has been applied to other problems in ride-sourcing (e.g., pricing, routing, trip dispatching). The readers are referred to \cite{liang2025machine} for a comprehensive review, while the following discussion focuses only on those most relevant to this paper. 
Given that vehicle rebalancing is a centralized control problem, a number of studies model it in the single-agent RL framework~\citep[e.g.,][]{oda2018movi,mao2020dispatch,jiao2021real}, with some adopting the graph neural network to capture the network structure in the ride-sourcing system~\citep{gammelli2021graph,gammelli2023graph}.
In addition to tackling rebalancing separately, some studies also jointly solve it with trip dispatching~\citep{gueriau2018samod,holler2019deep,jin2019coride}. 
However, the single-RL framework invariably suffers from the scalability issue. As the rebalancing decision is typically modeled as the number of vehicles moving from one local service to another, the dimension of action space increases rapidly with the number of zones defined in the system. 
To tackle this issue, recent studies have reformulated the rebalancing problem as multi-agent reinforcement learning (MARL)~\citep{busoniu2008comprehensive}. 
Instead of having a central controller decide all rebalancing flows, \citet{liu2021meta} models each zone as an agent that determines how many vacant vehicles it dispatches to neighboring zones. Differently, in \citet{lin2018efficient}, each vacant vehicle is modeled as an agent that selects its next search destination. 

Although MARL greatly reduces the dimension of the action space, it leads to another challenge, i.e., the communication and coordination among agents. A typical solution is to introduce a central controller that collects and broadcasts information to agents. Accordingly, each agent may update its policy based on the global information and local observations~\citep{foerster2016learning,lowe2017multi}. 

Mean-field reinforcement learning (MFRL) provides another promising path to resolve the scalability issue. MFRL has been used to model the searching behaviors of non-cooperative ride-sourcing drivers~\citep{shou2020optimal} and to decentralize the order dispatching problem in ride-sharing~\citep{li2019efficient}. 
MRFL has also been recently introduced to address the rebalancing problem~\citep{li2025repositioning,ling2025balancing,ye2025dynamic}.
Specifically, in \cite{steinberg2025mean}, the rebalancing problem is modeled as a contextual multi-armed bandit, which can be seen as a single-step RL problem, and optimized via mean-field Bayesian methods. 
Our previous work applied safe model-based MFRL to tackle the rebalancing problem, assuming unknown deterministic transition dynamics~\cite{jusup2024safe}. Differently, this paper incorporates stochastic transitions due to uncertainties in vehicle-rider matching and the demand process. 

As mentioned in \Cref{sec:introduction}, the equity issue has rarely been addressed in vehicle rebalancing, and it has been introduced only as an auxiliary reward \citep{wang2021data, sun2022optimizing}. To the best of our knowledge, our previous work \citep{jusup2024safe} is the first to explicitly express fairness as an accessibility guarantee and incorporate it into the MFRL framework as a constraint, leading to a safe MFRL problem.

\begin{longtable}{>{\footnotesize}p{4cm}| *{6}{>{\footnotesize}c}}
    \caption{Comparison of RL-based vehicle rebalancing studies across six methodological dimensions. Our work satisfies all.} 
    \label{tab:literature-review} \\
    \toprule
    Literature & Mean-field & Safe RL & Equity & Multi-agent & Model-based & Continuous \\ 
    \midrule
    \citep{mao2020dispatch, gammelli2021graph} & \xmark & \xmark & \xmark & \xmark & \xmark & \xmark \\ 
    \citep{yang2022learning, yu2019markov, gammelli2023graph} & \xmark & \xmark & \xmark & \xmark & \cmark & \xmark \\ 
    \citep{oda2018movi, jin2019coride, lin2018efficient, gueriau2018samod, liu2021meta, wen2017rebalancing, lee2025ride} & \xmark & \xmark & \xmark & \cmark & \xmark & \xmark \\ 
    \citep{holler2019deep} & \xmark & \xmark & \xmark & \cmark & \xmark & \cmark \\ 
    \citep{shou2020optimal} & \xmark & \xmark & \xmark & \cmark & \cmark & \xmark \\ 
    \citep{wang2021data, sun2022optimizing} & \xmark & \xmark & \cmark & \cmark & \xmark & \xmark \\  
    \citep{jiao2021real} & \xmark & \xmark & \xmark & \cmark & \cmark & \cmark \\ 
    \citep{shou2020reward, ling2025balancing, steinberg2025mean} & \cmark & \xmark & \xmark & \cmark & \xmark & \xmark \\
    \citep{li2025repositioning} & \cmark & \xmark & \xmark & \cmark & \cmark & \xmark \\ 
    \textbf{This paper}, \citep{jusup2024safe} & \cmark & \cmark & \cmark & \cmark & \cmark & \cmark \\ 
    \bottomrule
\end{longtable}

\subsection{Safe Reinforcement Learning}
Safe RL focuses on learning optimal policies while enforcing (safety) constraints during both training and deployment~\citep{garcia2015comprehensive,gu2022review}. A classical formulation treats safety as a constraint on expected cumulative cost, where the agent maximizes reward while satisfying one or more cost constraints per episode or in expectation~\citep{altman1999constrained}. Most research focuses on a single-agent RL safety~\citep{achiam2017constrained,moldovan2012safe,song2012efficient,gehring2013smart,berkenkamp2017safe,cheng2019end,alshiekh2018safe}, but lately, multi-agent RL safety has also gained increased attention.
For the cooperative problem MARL,~\citet{gu2021multi} propose MACPO and MAPPO-Lagrangian. The algorithms have the advantage of being model-free but come at the cost of MACPO being computationally expensive, while MAPPO-Lagrangian does not guarantee hard constraints.
\citet{lu2021decentralized} introduce Dec-PG to solve the decentralized learning problem by sharing weights between neighboring agents in a consensus network.
CMIX~\citep{liu2021cmix} extends QMIX~\citep{rashid2020monotonic} to focus on population-based constraints. They consider a centralized-learning decentralized-execution framework to satisfy average and peak constraints defined for the population. The proposed approach does not scale well because it relies on joint state and action spaces.
\citet{elsayed2021safe} utilize shielding to correct unsafe actions, but their centralized approach suffers from scalability issues. Factorized shielding improves scaling at the expense of monitoring only a subset of the state or action space.
In the MFC setting, \citet{mondal2022mean} introduces constraints by defining a cost function and a threshold that the discounted sum of costs can not exceed. 
We take the approach discussed in \citet{jusup2024safe}, which restricts the set of feasible mean-field distributions at every step. Their approach addresses the scalability issue and allows for more specific control over constraints and safe population distribution.

\subsection{Mean-Field Control and Mean-Field Reinforcement Learning} \label{sec:mfc-and-mfrl-review}
We refer to MFC as the setting in which a closed-form transition model is available (e.g., provided by a domain expert), and reserve MFRL for scenarios where the transition dynamics must be learned through interaction with the environment.
\citet{gast2012mean} are the first to analyze MFC as a Mean-Field Markov Decision Process (MF-MDP) and show that the optimal reward converges to the solution of a continuous Hamilton-Jacobi-Bellman equation under some conditions.
\citet{Bauerle2021MeanProcesses} formulate MFC as an MF-MDP where the distribution is defined as an empirical measure of the agents' states. They show the existence of an $\epsilon$-optimal policy under some conditions.
\citet{Motte2019Mean-fieldControls} show the existence of $\epsilon$-optimal policies for more general MF-MDPs with continuous state and action spaces. 
\citet{carmona2019model} consider a limiting distribution of continuous agents' states to define MFC as an MF-MDP to show an optimal policy exists. Our algorithm implementation was inspired by their discussion of the discretization strategy for MFQ-learning.   
\citet{chen2021pessimism} discuss settings where the interaction with the environment during training can be expensive, prohibitive, or unethical. They introduce an offline MFRL algorithm, SAFARI, and analyze its sub-optimality gap.
\citet{Gu2019DynamicMFCs,Gu2021Mean-FieldAnalysis} show that model-free kernel-based Q-learning has a linear convergence rate for MFC. They also establish the MFC approximation of cooperative MARL with $L$ agents as  $\bigO(\nicefrac{1}{\sqrt{L}})$. \citet{hu2023graphon} devise the same approximation error for graphon MFC, a limiting object of large dense graphs. 
\citet{pasztor2021efficient} show a sublinear cumulative regret for a model-based MFRL algorithm, M\textsuperscript{3}-UCRL.
Our work differs in that it also incorporates an exogenous stochastic process---demand---into the system dynamics. Establishing theoretical bounds with respect to the complexity of such exogenous processes would be highly beneficial for the transportation community, yet it remains an open challenge.
Closest to the setting analyzed in this manuscript, our previous study~\citep{jusup2024safe} extends \citet{pasztor2021efficient} with safety constraints. Safe-M\textsuperscript{3}-UCRL optimizes the reward and learns underlying dynamics while satisfying constraints throughout the execution. What distinguishes the present work is that we impose the accessibility constraint only on available vehicles and evaluate it exclusively over residential zones, rather than on the entire fleet across all zones, including green areas and rivers.

\section{Model Setup} \label{sec:model-setup}
We consider a single ride-sourcing platform that provides on-demand trips within a certain service region and operation time horizon using a large fleet of vehicles. Particularly, vehicles are assumed to be homogeneous and cooperative, thus fully compliant with the platform's dispatches.
In terms of the system dynamics, the following assumptions are introduced. First, the operation time horizon is discretized into fixed intervals, and all passenger and rebalancing trips are completed within one interval. The service region, however, remains continuous, and thus vehicle locations at the beginning of each time interval are presented by their instantaneous coordinates. We further assume the platform can observe all vehicles' locations in real time and dispatch them to specific coordinates. On the other hand, the platform only has partial knowledge about the passenger demand and dispatches nearby vacant vehicles to serve emerging ride requests. Therefore, to match vehicle supply with passenger demand, the platform must reposition vacant vehicles in anticipation of future demand prediction based on historical data. 
In addition to maximizing the number of served passengers, the platform aims to achieve equity by enforcing a minimum level of service accessibility over space. Since each passenger can only be served by nearby vacant vehicles, the accessibility is measured by the spatial distribution of vacant vehicles.

The remainder of this section is organized as follows. \Cref{sec:C-MF-MDP} formulates the rebalancing problem as a constrained mean-field Markov decision process; \Cref{sec:transition} specifies the individual state and mean-field transitions dictated by the cruising, matching, and repositioning processes; \Cref{sec:reward} defines the reward that signals the rebalancing decisions; and \Cref{sec:accessibility-constraint} formally defines the service accessibility constraint. 

\subsection{Constrained Mean-Field Markov Decision Process}\label{sec:C-MF-MDP}
We formulate the rebalancing problem for a single representative vehicle as a \textit{constrained mean-field Markov decision process} (\cmfmdp) represented by a tuple:
\begin{align}
    \mdp=(\s, \A, \pps, \e, f, r, h, \mu_0, C), 
\end{align}
with key objects defined as follows: 
\begin{itemize}
    \item State space $\s \subseteq \R^2$ refers to the two-dimensional service region, and state $\st \in \s$ denotes the location (coordinates) of the vehicle at time $t$.
    
    \item Action space $\A:= [0,1] \times \R^2$ represents the set of all possible actions. Action $\at=(\pt,\ut)$ consists of the repositioning odds $\pt \in [0,1]$ and the corresponding movement $\ut\in\R^2$ at time $t$. Having predetermined unbiased repositioning odds is an implicit way of incorporating fairness toward drivers who lose profit during repositioning. \looseness=-1
    
    \item We use $\bbs$, $\mms$, and $\pps$ to represent Borel sets, a set of Borel measures, and a set of absolutely continuous probability measures over $\s$, respectively.

    \item Mean-field distribution $\mut\in \pps$ is defined as the limiting vehicle distribution at time $t$ with fleet size $L$ increasing to infinity, i.e., 
    \begin{align}
        \mut(\ds') = \lim_{L \to \infty} \mut^L(\ds'),
    \end{align}
    where $\mut^L$ is the empirical mean-field distribution defined as:
    \begin{align}
        \mut^L(\ds') = \frac{1}{L} \sum_{i=1}^L \I(\st^{(i)}\in \ds'),
    \end{align}
    where $\I(\cdot)$ is the indicator function, $\st^{(i)}$ denotes the state of vehicle $i$ and $\ds'$ is an inifinitesimal interval around state $s'$. \looseness=-1
    
    In the remainder of the paper, we will use notations $\mu(\ds)$ for probability measures/distributions and $\mu(s)$ for associated probability densities. Note that the assumption of absolute continuity of measures $\mu(\ds)$ with respect to the Lebesgue measure implies that associated densities $\mu(s)$ always exist. 
    The mean-field distribution of vehicles at time $t=0$, i.e., $\mu_0\in \pps$, is also referred to as the initial mean-field distribution.

    \item Exogenous processes $\e$ influence decisions and transitions, but their future realizations are not impacted by our actions. Concretely, we use $\e = \mms \times \pps^\s$ to represent the knowledge of the demand pattern $\etat=(\deltat,\Phit) \in \e$ at time $t$ consisting of a ride demand measure $\deltat \in \mms$ and the trip origin-destination transition $\Phit \in \pps^\s$. 
    Given a finite demand rate $d_t\in \R_{\geq 0}$, a demand distribution $\odeltat\in \pps$, and a finite fleet size $L \in \R_+$, the demand measure is defined as $\deltat:=(\sfrac{d_t}{L})\odeltat \in\mms$. 
    
    The origin-demand transition $\Phit$, which maps from trip origin to a probability distribution of the destinations, is modeled by a Markov kernel $\Phit: \bbs \times \s \to [0, 1]$.
    We write $\Phit \in \pps^\s$ to indicate that every origin $s$ is mapped to a probability distribution over $\s$, i.e., $\int_{s'\in\s}\Phit(\ds', s)=1$ for every $s\in\s$.

    \item State transition function $f:\s\times \pps \times \e \times \A \to \pps$ maps from the current state $\st\in\s$, the mean-field distribution $\mut \in \pps$ and the demand pattern $\etat \in \e$ to the probability distribution of next state $\stn\sim\pps$ given action $a\in\A$.

    \item Time-dependent policy profile $\bpi \eq(\pi_{0},\ldots,\pi_{T-1}) \in \bPi$ is a sequence of policies, where each policy $\pit: \s \times \pps \times \mms \to \A$ maps from the vehicle's state $s_t$ and the supply-demand relationship summarized by $(\mut, \deltat) \in \pps \times \mms$ to a deterministic action $\at \in \A$.  We use  $\Pit$, for $t=0, 1, \ldots, T-1$ and $\bPi = \Pi_0 \times \Pi_1 \times \cdots \times \Pi_{T-1}$ to denote sets of admissible policies and admissible policy profiles, respectively.

    \item Reward function $r:\pps\times\mms\times\Pi \to [0,1]$ evaluates the immediate reward of policy $\pit\in\Pit$ learned by the representative vehicle given the distribution of vehicles $\mut\in\pps$, and the demand measure $\deltat\in\mms$.
    Note that here, we use the population-level reward, commonly called the lifted reward, rather than the vehicle-level reward. Given the assumption of homogeneous vehicles, it can be interpreted as the average reward received by vehicles in the fleet. 

    \item Accessibility constraint is defined by setting a lower bound $C\in\R$ on an accessibility function $h(\cdot)$, i.e., $h(\cdot) \geq C$, where a function $h: \pps\times\Pi \to \R$ measures service accessibility given the distribution of vehicles $\mut$ and a policy $\pit$.   
\end{itemize}

\subsection{State and Mean-Field Transitions}\label{sec:transition}
\begin{figure}[htb!]
    \centering
    \includegraphics[width=0.7\textwidth, trim=3cm 5.5cm 10cm 5cm, clip]{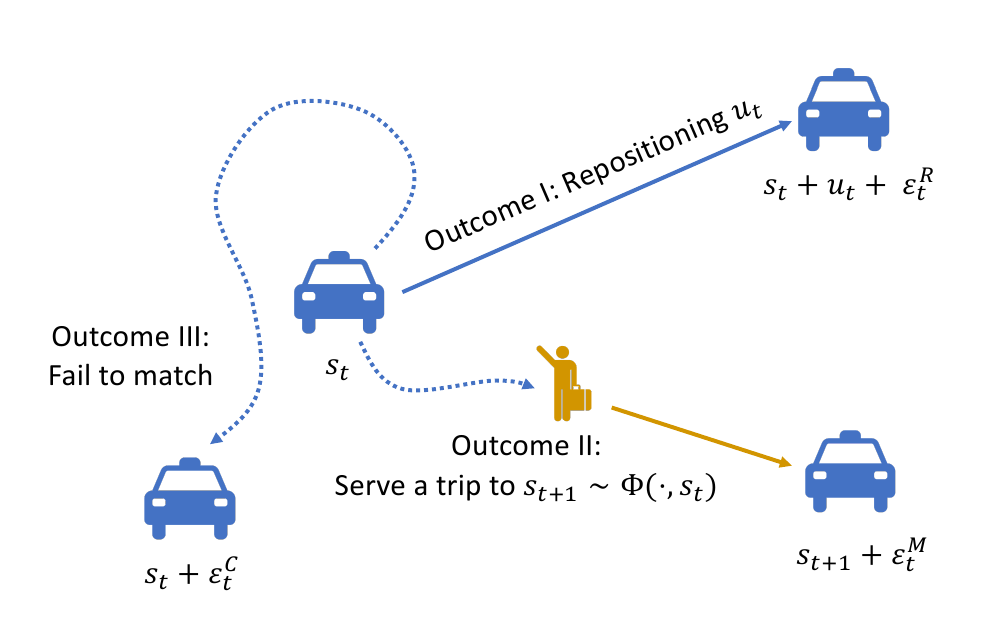}
    \caption{Illustration of the representative vehicle state transitions.}
    \label{fig:state_transition}
\end{figure}

We decompose the state transition function influenced by stochastic exogenous processes (e.g., rider preferences) into three components depending on whether the controller repositions the vehicle in anticipation of future demand, the vehicle gets matched and is fulfilling a ride in the given time step, or cruises if it fails to match with a rider.
\Cref{fig:state_transition} illustrates the described process within a time interval $[t, t+1)$. At the beginning $t$ of the interval, a controller uses a policy $\pit$ to assign a repositioning probability $\pt$ and movement $\ut$ to a vehicle at the location $\st$. In other words, with probability $\pt$, a vacant vehicle continues its next step $t+1$ around the location $\st+\ut$ to ensure accessibility and/or in anticipation of future demand. There are two possible outcomes if the vehicle does not reposition: (i) with probability $\mt$, the vehicle matches with a rider and ends the trip according to the rider's preferences, i.e., near the location determined by the origin-destination transition $\Phit(\cdot,\st)$, and otherwise (ii) the vehicle fails to find a rider after cruising over the entire interval and stays at a location nearby the current location $\st$. 

Accordingly, the state transition function is specified as:
\begin{align} \label{eq:state_transition}
    f(\st,\mut,\etat,\at) = 
    \begin{cases}
        \begin{aligned}
            &\st + \ut, &&\text{ w.p. } \pt \quad &&\text{(R: repositioning)}\\
            &\stn\sim \Phit(\cdot, \st), &&\text{ w.p. } (1-\pt)\mt \quad &&\text{(M: matched)} \\
            &\st, &&\text{ w.p. } (1-\pt)(1-\mt). \quad &&\text{(C: cruising)}
        \end{aligned}
    \end{cases}
\end{align}

In \Cref{eq:state_transition}, the repositioning probability $\pt$ and vehicle movement $\ut$ are given by the policy $\pit$, i.e., 
\begin{align}
    (\pt, \ut) = \pit(\st,\mut,\deltat).
\end{align}
Once the repositioning actions are determined, the matching probability $\mt$ depends on the available vehicles $\muta$ and ride demand $\deltat$ at time $t$. Specifically, it is specified by some \textit{matching process} $M:\s\times\mms\times\mms\to [0,1]$ as:
\begin{align}\label{eq:matching_probability}
    \mt = M(\st, \muta,\deltat),
\end{align}
where the measure of available vehicles $\muta\in \mms$ is given by: 
\begin{align}
    \muta(\ds) &=(1-\pt)\mut(\ds).
\end{align}

We further assume that each vehicle experiences independent idiosyncratic noise, which diminishes in aggregate as the fleet size increases, and thus vanishes in the limit.
Formally, we introduce it as Gaussian noise associated with each case in~\Cref{eq:state_transition}:
\begin{align} \label{eq:noise}
    \epsilont = 
    \begin{cases}
        \epsilontr & \text{(R: repositioning)}\\
        \epsilontm & \text{(M: matched)}\\
        \epsilontc, & \text{(C: cruising)}\\
    \end{cases}
\end{align}
with zero mean and known variances to represent the randomness of the vehicle's ending location. Specifically, to reflect the real practice, we set a much larger variance to $\epsilontr$ and $\epsilontc$ but a negligible variance to $\epsilontm$ (i.e., the trip drop-off location is nearly certain). 
Accordingly, the next state $\stn$ is sampled from the stochastic state transition $f$ and noisy ending location, i.e.,
\begin{align}\label{eq:next-state}
    \stn\sim f(\st,\mut,\etat,\at) + \epsilont.
\end{align}

The fleet behavior can be seen as the aggregate behavior of individual vehicles, i.e., the mean-field distribution $\mut$ can be decomposed into three components: (i) repositioning vehicles measure $\mutr\in\mms$, (ii) matched vehicles measure $\mutm\in\mms$, and (iii) cruising vehicles measure $\mutc\in\mms$:
\begin{align}\label{eq:mf_decomposition}
    \mut(\ds) \eq \mutr(\ds) &+ \mutm(\ds) + \mutc(\ds),
\end{align}
where 
\begin{subequations} 
\begin{align} 
    \mutr(\ds) &\eq \pt\mut(\ds), \label{eq:mf_repositioning} \\ 
    \mutm(\ds) &\eq \mt(1-\pt)\mut(\ds), \label{eq:mf_matched} \\ 
    \mutc(\ds) &\eq (1-\mt)(1-\pt)\mut(\ds).\label{eq:mf_cruising} 
\end{align}
\end{subequations}
Hence, the measure of available vehicles can also be expressed as: 
\begin{align}
    \muta(\ds) = \mut(\ds) - \mutr(\ds) = \mutm(\ds) + \mutc(\ds).
\end{align}

Since the state transition $f$ describes the transition for the representative vehicle, it is applicable to every vehicle in the fleet. Therefore, $f$ induces the transition dynamics of the mean-field distribution.
Following \citet{jusup2024safe}, we defined the mean-field transition by: 
\begin{equation} 
    \mutn(\ds') \eq\int_\sis \p[\stn \in \ds' | \st=s]\mut(\ds),
\end{equation}
which can in our case be obtained as a sum of transitions of $\mut$ components in \Crefrange{eq:mf_repositioning}{eq:mf_cruising} as follows: 
\begin{align} \label{eq:mf-transition}
\begin{split}
    \mutn(\ds') = &\int_\sis \p[\st + \ut + \epsilontr \in \ds' | \st=s]\mutr(\ds) \\
    &+ \int_\sis \p[\Phit(\cdot, \st) + \epsilontm \in \ds' | \st=s]\mutm(\ds) \\
    &+ \int_\sis \p[\st + \epsilontc \in \ds' | \st=s]\mutc(\ds).
\end{split}
\end{align}

Note that the mean-field transition is a function of the mean-field distribution $\mut$, exogenous rider preferences $\Phit$, state transition $f$, and a policy $\pit$ because of the assumptions of homogeneous vehicles. 
Hence, the mean-field transition, denoted by $U(\cdot)$, can be written in a more compact form as: 
\begin{align} \label{eq:mf_transition}
    \mutn = U(\mut, \etat, \pit, f).
\end{align}

\subsection{Reward Specification}\label{sec:reward}

As per \Cref{sec:C-MF-MDP}, we consider a population-level reward shared by cooperative vehicles defined on the mean-field distribution $\mut$ and policy $\pit$. 
Since the ultimate goal of vehicle rebalancing is to maximize the utilization rate (i.e., fleet matching rate), we first define the unnormalized reward as: 
\begin{align}\label{eq:unnormalized-reward}
    R(\mut, \deltat, \pit) = \mutm(\s) - \djs(\omuta  || \odeltat),
\end{align}
where $\mutm(\s) = \int_{\sis} \mutm(\ds) \in [0,1]$ represents a fraction of matched vehicles, $\omuta(\ds') = \sfrac{\muta(\ds')}{\muta(\s)} \in \pps$ represents the distribution of available vehicles, $\odeltat(\ds') = \sfrac{\deltat(\ds')}{\deltat(\s)} = (\sfrac{L}{d_t})\deltat(\ds') \in \pps$ represents the distribution of demand and $\djs:\pps\times\pps \to [0,1]$ is Jensen-Shannon (JS) divergence that measures the similarity between the distributions of available vehicles $\omuta$ and demand $\odeltat$. 
Note that the two terms are, respectively, ex-post and ex-ante performance metrics of vehicle rebalancing actions. They are both necessary to stabilize and efficiently navigate the learning process. 

The reward \Cref{eq:unnormalized-reward} is then normalized to provide a more intuitive 0 to 1 utilization rate scale: 
\begin{align} \label{eq:reward}
    r(\mut,\deltat,\pit) = \frac{1}{2} \Big(R(\mut, \deltat, \pit) + 1\Big).
\end{align}

\subsection{Accessibility Constraint} \label{sec:accessibility-constraint}
While the overall objective is to maximize the fleet utilization rate, we also want to ensure equitable service access. 
Directly optimizing multiple business-oriented accessibility metrics (e.g., those introduced in \Cref{sec:metrics}) during training is possible but risks biasing the policy toward specific metrics. Instead, we define a general accessibility constraint that promotes a sufficient spatial spread of available vehicles across the service region. This spread acts as a proxy for accessibility: by maintaining vehicle availability throughout the network, the policy is more likely to achieve high performance across diverse accessibility metrics during evaluation. The spread is naturally quantified using the $\varepsilon$-smoothed weighted differential entropy, which captures both coverage and concentration of available vehicles:
\begin{align} \label{eq:weighted-differential-entropy}
    \hw(\mut, \pit) := -\int_\sis w(s)\log(\omuta(s) + \varepsilon)ds,
\end{align}
where the weight function $w:\s\to \R_{\geq 0}$ represents the importance of each location, while $\varepsilon > 0$ ensures that the entropy is bounded. An important use case is to measure the spread of vehicles across the service region while ignoring non-operating locations like rivers and green spaces by assigning them zero weights.

We then introduce $C\in\R$ as a desired lower bound of accessibility, and accordingly, the accessibility constraint is given by: \looseness=-1
\begin{align} \label{eq:weighted-accessibility-constraint}
    \hw(\mut,\pit) {}\geq{} C.
\end{align}

Intuitively, entropy quantifies how concentrated the distribution of available vehicles is, reaching its maximum when the distribution is uniform. Imposing a lower bound on entropy constrains how concentrated the fleet may become, thereby enforcing a minimum level of spatial accessibility. Consequently, the model optimizes the objective in \Cref{eq:mfc-objective} over all feasible vehicle distributions, balancing two goals: (i) keeping available vehicles close to riders, and (ii) maintaining accessibility by preventing excessive concentration around demand hotspots.

\section{MFRL and MFC for Vehicle Rebalancing} \label{sec:mfc-and-mfrl}
With all elements specified in the previous section, we are now ready to formalize the vehicle rebalancing problem as the optimization problem faced by the representative vehicle. Given the initial distribution $\mu_0$, the optimal vehicle rebalancing policy profile $\bpis$ is a solution to the following \cmfmdp:

\begin{subequations}\label{eq:mfc}
\begin{align}
    \argmax_{\bpi \in \bPi} {}\; & \E \Bigg [ \sum_{t=0}^{T-1} r(\mut, \deltat, \pit) \Big\rvert \mu_{0}\Bigg ] \label{eq:mfc-objective} \\
    \text{s.t.} \quad & \at \eq\pitsmr \\
    & \stn \sim{} f(\st, \mut, \etat, \at) + \epsilont \label{eq:mfc-transition}\\
    &\mutn \eq\Umpf \label{eq:mfc-mf-transition} \\
    &\hw(\mutn, \pitn) \geq{} C. \label{eq:mfc-safety}
\end{align}
\end{subequations}

However, the state transition $f$ is rarely known a priori in practice. In this study, we propose two approaches to tackle this issue. The first is constructing a closed-form approximation $\hf$ with domain knowledge, thus turning \Cref{eq:mfc} into a mean-field control (MFC) problem. The second approach is to learn an approximation $\tf$ via episodic interactions with the environment, leading to a model-based mean-field reinforcement learning (MFRL) problem. \looseness=-1

In what follows, Sections~\ref{sec:ot_approximation} and \ref{sec:learning_matching_process} explain how to approximate the matching process, the key unknown component in $f$, by optimal transport and a model-based learning protocol, and \Cref{sec:log-barrier-method} introduces the log-barrier method adopted to tackle the accessibility constraint during the optimization.

\subsection{Optimal Transport Approximation of Matching Process in MFC} \label{sec:ot_approximation}
Recall that the state transition $f$ specified in \Cref{eq:state_transition} depends on two processes. One is the rider delivery process solely driven by the origin-destination transition $\Phit$ that can be reasonably estimated from historical data. The other is the matching process $M$ that is determined by both ride demand $\deltat$ and available vehicle supply $\muta$. 
Instead of characterizing the sophisticated vehicle-rider matching mechanism~\citep[e.g.,][]{zhang2023ride}, we propose to approximate $M$ as an \textit{optimal transport} (OT) between the demand and supply distributions. 

\begin{definition}[Kantorovich's OT formulation]
    Let $X$ and $Y$ be two separable metric spaces and let $c:X\times Y\to \R_{\geq0}$ be a Borel-measurable function. Given probability measures $\tau$ on $X$ and $\nu$ on $Y$, the optimal transport is defined as a probability measure $\gamma$ on $X\times Y$ such that:
    \begin{align} 
        \gamma  = \arginf_{\gamma\in\Gamma(\tau,\nu)}\int_{X\times Y}c(x,y)d\gamma(x,y),
    \end{align}
    where $\Gamma(\tau,\nu)$ is the set of all couplings of $\tau$ and $\nu$ (i.e., a joint probability measures with marginals $\tau$ and $\nu$). 
\end{definition}

In short, $\gamma$ offers a ``transport plan'' that optimally moves probability mass from $\tau$ to $\nu$. To fit in the context of vehicle-rider matching, we set $X=Y=\s$, specify $\tau = \omuta, \nu = \odeltat$, and define the cost function $c$ as the Euclidean distance $d:\s\times\s\to\R_{\geq0}$.
To align with real practice, we further impose a matching radius $r>0$ that dictates the maximum distance between matched riders and available vehicles.

We then use the disintegration theorem \citep[Theorem~33.3 in][]{Bill86,pachl1978disintegration} to define the approximated matching process $\hM: \s\times\mms\times\mms\to [0,1]$ with optimal transport $\gamma$ as:
\begin{align}\label{eq:matching_process_approx}
    \hM(s, \muta, \deltat) = \int_{s'\in N_r(s)}  \gamma(s, \ds'),
\end{align} 
where $N_r(s)=\{s'\in\s:d(s,s')<r\}$ denotes the neighborhood of $s$ with radius $r$. 

In other words, the matching outcome of a vehicle at location $s$ is dictated by the optimal transport $\gamma$ and the distance threshold $r$. 
We then plug the matching process approximation $\hM$ into the state transition function in \Cref{eq:state_transition} to get an approximated state transition function $\hf$. We then use MFC to solve \Cref{eq:mfc} under transitions $\hf$ to obtain a policy profile $\bpisot$.

\subsection{Learning of Matching Process in MFRL} \label{sec:learning_matching_process}
Since the matching process $M$ is a complex stochastic process, it is not guaranteed that the OT will approximate all its realizations well. To better capture the underlying complexity, MFRL learns to approximate the matching process $M$ through the \textit{episodic} interactions of the representative vehicles with the environment. 
Each episode $n=1,2,\dots,N$ consists of $T$ discrete time steps $t=0,1,\dots,T-1$. In the context of ride-sourcing, the episodes represent the operational units (e.g., working days) while the time steps define the short-term periods during which the rebalancing operations are performed. 

To learn the matching process $M$, we collect trajectories $\D_n = \{((\snt, \muant, \deltant), \bnt)\}_{t=0}^{T-1}$ of the representative vehicle in each episode $n$, which include its own state $\snt$, measures of the available vehicles $\muant$ and ride demand measure $\deltant$, as well as a binary label $\bnt\in\{0,1\}$ that indicates whether the representative vehicle is matched with a rider at time $t$. 
The trajectories up to episode $n-1$, i.e., $\cup_{i=1}^{n-1} \D_i$, are used to train a binary classifier $\tM_{n-1}:\s\times\mms\times\mms\rightarrow [0,1]$ that predicts the matching probability given the supply-demand relationship. The classifier is then plugged into \Cref{eq:state_transition} to derive the approximate state transition function $\tf_{n-1}$. Solving \Cref{eq:mfc} with $\tf_{n-1}$ yields the optimal policy profile $\bpins$, which is deployed in episode $n$.

The MFRL learning protocol is summarized in the \Cref{alg:learning_protocol}. Notice that MFC performs a single execution of Line 2. Hence, it is computationally more efficient and appealing for applications with limited time and/or budget. On the other hand, MFRL can learn the matching process directly from the environment rather than using generic and possibly biased expert models, which makes it advantageous when succinct data is obtainable.

\begin{algorithm*}[htb]
\caption{Model-Based MFRL Learning Protocol}
\label{alg:learning_protocol}
\KwIn{Set of admissible policy profiles $\bPi$, initial mean-field distribution $\mu_0$, matching process prior $\tM_0$, reward $r(\cdot)$, accessibility function $\hw(\cdot)$, exogenous processes estimation $\etat = (\deltat, \Phit)$, accessibility lower-bound $C$, log-barrier hyperparameter $\lambda$, number of episodes $N$, number of steps $T$}
\For{$n \leftarrow 1$ \KwTo $N$}{
  Optimize \cmfmdp $\mdp=(\s, \A, \pps, \e, f, r, h, \mu_0, C)$ using log-barrier method in \Cref{eq:log-barrier} over the admissible policy profiles $\bPi$, under state transition function $f=\tf_{n-1}$ induced by learned matching process $\tM_{n-1}$ and subject to the accessibility constraint $\hw(\cdot)\geq C$ to obtain the policy profile $\bpi_n^*$\;
  
  Execute the obtained policy profile $\bpi_n^*$ and collect the trajectory $\D_n = \{((\snt, \muant, \deltant), \bnt)\}_{t=0}^{T-1}$ observed by the representative vehicle\;
  
  Train the matching process approximation $\tMn$ using historical trajectories $\cup_{i=1}^{n} \D_i$\;
}
\KwOut{$\bpi_N^* \eq(\pi_{N,0}^*,\ldots,\pi_{N,T-1}^*)$}
\end{algorithm*}

\subsection{Log-Barrier Method} \label{sec:log-barrier-method}
The accessibility constraint makes it particularly challenging to solve \Cref{eq:mfc}. To address this issue, we apply the \textit{log-barrier method}~\citep{wright1992interior} to turn \Cref{eq:mfc} into an unconstrained optimization problem compatible with standard solvers, as elaborated in \citet{jusup2024safe}. The reformulated problem becomes:
\begin{subequations}\label{eq:log-barrier}
\begin{align}
    \argmax_{\bpi \in \bPi} {}\; & \E \Bigg [ \sum_{t=0}^{T-1} r(\mut, \deltat, \pit) + \lambda\log(\hw(\mutn, \pitn) - C)
    \Big\rvert \mu_{0}\Bigg ] \\ \label{eq:log-barrier-objective}
    \text{s.t.} \quad & \at \eq\pitsmr \\
    & \stn \sim{} f(\st, \mut, \etat, \at) + \epsilont \label{eq:log-barrier-transition}\\
    & \mutn \eq\Umpf, \label{eq:log-barrier-mf-transition}
\end{align}
\end{subequations}
where $\lambda > 0$ is a hyperparameter that balances the reward $r(\cdot)$ and the accessibility constraint slack $\hw(\cdot) - C$. In \Cref{apx:impact-of-lambda}, we discuss how the choice of $\lambda$ affects the optimization.

The above reformulation restricts the domain on which the objective function is defined only to values that satisfy the constraint, i.e., the term $\log(\hw(\mut, \pit) - C)$ is undefined if $\hw(\mut, \pit) \leq C$. Therefore, the log-barrier method guarantees that $\hw(\mut, \pit) > C$ for $t > 0$, which is empirically confirmed in \Cref{apx:constraint-satisfaction}. However, if the accessibility constraint is required to be satisfied at the initial step $t=0$, special care must be taken during initialization. We elaborate on efficient initialization strategies in \Cref{sec:learning-protocol-setup}.
Since the reformulation in \Cref{eq:log-barrier} relaxes the hard constraint in \Cref{eq:mfc-safety} into a soft constraint embedded in the objective function, the two formulations are not guaranteed to converge to the same policy. As we will see in \Cref{sec:experiments}, the log-barrier method leads to high-performance \textit{conservative} policies, i.e., $\hw(\cdot) \gg C$.

\section{Experiments} \label{sec:experiments}
The experiments presented in this section are based on a large dataset of taxi trajectories collected over five weeks in Shenzhen, China, from January \nth{18}, 2016 to September \nth{25}, 2016. The dataset was provided by a local transportation authority and is therefore not publicly available. Nevertheless, \citet{nie2017can} offers a detailed statistical analysis of the data, which can be used to reconstruct a dataset with similar statistical properties for reproducibility.
All the experiments were run on a cluster with Intel Xeon Gold 511 CPU, Nvidia GeForce RTX 3090 GPU, and 64 GB RAM. The code is released in \citet{jusup_2025_14991605}. \looseness=-1

\subsection{Simulation Environment} \label{sec:simulation-environment}
The proposed vehicle rebalancing algorithms are trained and evaluated on a simulator with continuous states and actions and discrete time steps. Experiments cover a six-hour service period from 16:00 to 22:00, divided into 20-minute intervals ($T=18$ time steps). As discussed in \Cref{apx:demand-variability}, we focus on evening peak hours, which exhibit notable---though still limited---spatiotemporal demand variability (see \Cref{fig:demand_variability}) and is of primary interest to service providers. The 20-minute interval is chosen because it covers nearly 90\% of rides in the dataset, with median and mean durations of approximately 9.5 and 11.5 minutes, respectively.
We assume all trips, including rider and rebalancing trips, finish within a single time step, which is a common assumption in the literature~\citep[e.g.,][]{zhang2023ride,mao2020dispatch,wang2021data,sun2022optimizing,ling2025balancing}. Riders not matched within a time step leave the ride-sourcing system. These assumptions can be relaxed to capture longer trips and congestion effects during rush hours~\citep[see e.g.,][]{zhang2023ride}.
We restrict our analysis to a square region that covers the city center, with coordinates spanning from 114.015 to 114.14 degrees longitude and from 22.5 to 22.625 degrees latitude. 
Since the main focus of this study is fleet management, we assume exogenous demand estimates are available for each time step.

\subsubsection{Demand Pattern}\label{sec:demand_pattern}
\begin{figure*}[htb!]
\centering
\includegraphics[width=0.5\textwidth]{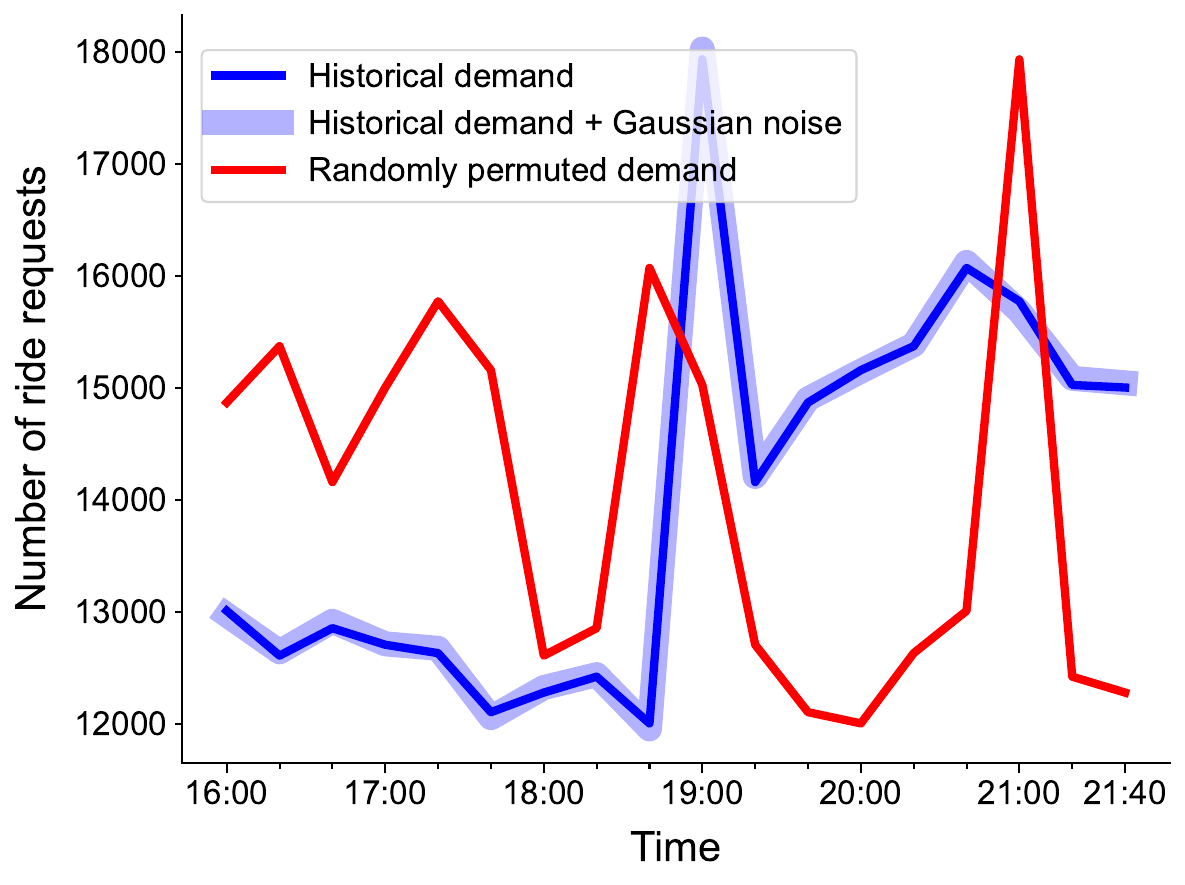} 
\caption{Aggregate demand rate over study period 16:00-22:00, specifically, the historical demand, the historical demand perturbed with Gaussian noise, and the randomly permuted demand.} \label{fig:demand_patterns}
\end{figure*}

\begin{figure*}[htb!]
\centering
\begin{subfigure}[t]{0.4\textwidth}
    \centering
    \includegraphics[width=\textwidth]{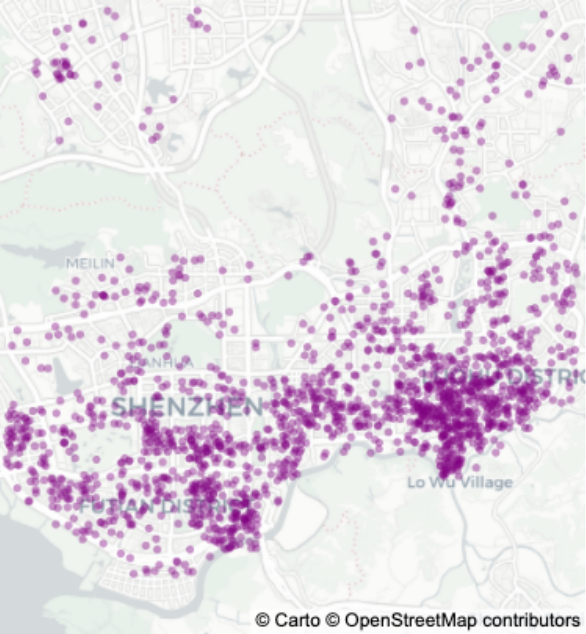}
    \captionsetup{justification=centering}
    \caption{Historical demand}
    \label{fig:requests_original}
\end{subfigure}
\hspace{0.2cm}
\begin{subfigure}[t]{0.4\textwidth}
    \centering
    \includegraphics[width=\textwidth]{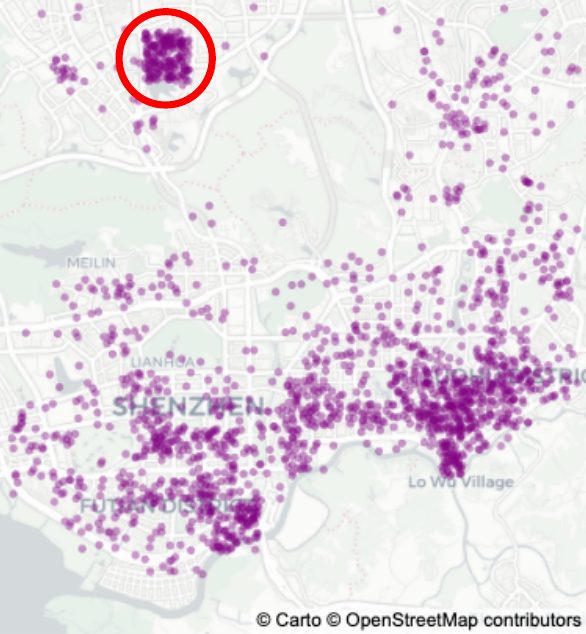}
    \captionsetup{justification=centering}
    \caption{Hypothetical demand shock}
    \label{fig:requests_shock}
\end{subfigure}
\caption{Demand distribution in the interval 20:00-20:20: (a) Pickups observed in historical data, and (b) with a hypothetical demand shock in a circled area consisting of four zones. }
\label{fig:requests_distributions}
\end{figure*}

Four demand patterns $\etat=(\deltat,\Phit) \in \e$ are constructed for model training and evaluation based on both real data and hypothetical scenarios. 
We analyze the performance under the demand patterns presented in Figures~\ref{fig:demand_patterns} and \ref{fig:requests_distributions}:
\begin{itemize}
    \item \textbf{Historical demand:} Demand pattern estimated from the real-world historical data. The proposed algorithms are trained on this demand pattern.
    \item \textbf{Historical demand + Gaussian noise:} A small Gaussian noise is added to the demand rate at each step. It is used to test the algorithms' robustness to tiny changes in demand magnitude.
    \item \textbf{Randomly permuted demand:} A random temporal permutation of demand rates over the study period. It is used to test the algorithms' robustness to significant changes in demand temporal patterns unseen during the training (see \Cref{fig:demand_patterns}). 
    \item \textbf{Hypothetical demand shock:} A demand shock is manually added to the northeastern corner of the study region between 20:00 and 21:00 (see \Cref{fig:requests_shock}). It is used to test the algorithms' robustness to temporary large events (e.g., a concert or a sports game). 

\end{itemize}

\subsubsection{State, Action and Probability Spaces Representation} \label{sec:state-space-representation}
We assume vehicles can reach any location using movements $\ut \in [-1,1]$ by rescaling the service region into a two-dimensional unit square, which yields the state space $\s=[0,1]^2$. 
To represent (probability) measures, we discretize the service region into a $25\times25$ grid where each zone is around $550\times 550$ meters. Note that the state space remains continuous.
The measure of each zone is then assigned to its center. Besides, the weight function $w(\cdot)$ used in \Cref{eq:weighted-differential-entropy} is specified according to whether or not each zone is operational for ride-sourcing vehicles. For instance, the zones corresponding to rivers and green spaces have weight $w=0$ because they are not physically accessible by vehicles. In total, 154 out of 625 zones are non-operational. Practitioners can further adjust $w(\cdot)$ to reflect business-specific priorities.

\subsubsection{Vehicle Initialization and Movements}\label{sec:vehicle_initial_movement}

\begin{figure*}[t!]
\centering
\begin{subfigure}[t]{0.16\textwidth}
    \centering
    \includegraphics[width=0.98\textwidth]{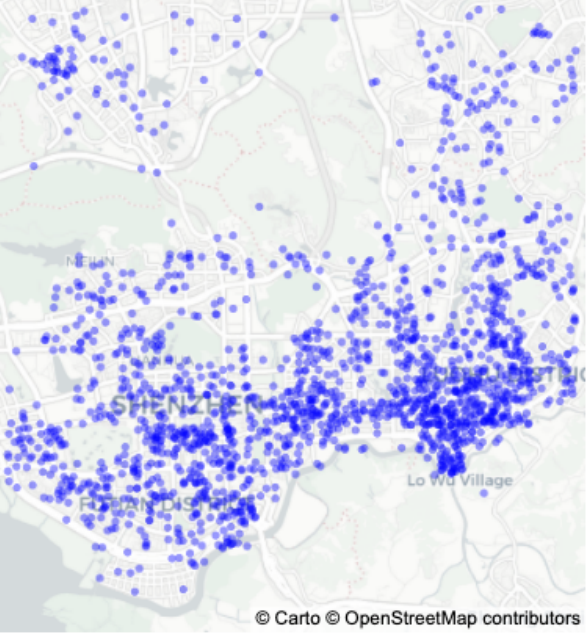}
    \captionsetup{justification=centering}
    \caption{$\alpha=0$}
\end{subfigure}
\hfill
\begin{subfigure}[t]{0.16\textwidth}
    \centering
    \includegraphics[width=0.98\textwidth]{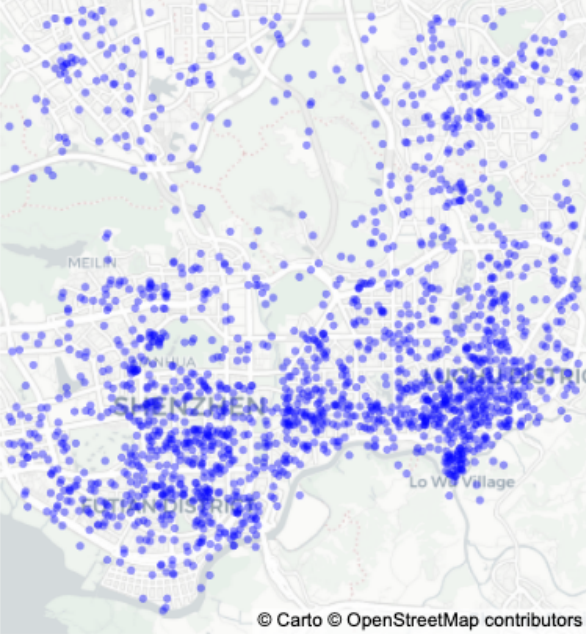}
    \captionsetup{justification=centering}
    \caption{$\alpha=0.2$}
\end{subfigure}
\hfill
\begin{subfigure}[t]{0.16\textwidth}
    \centering
    \includegraphics[width=0.98\textwidth]{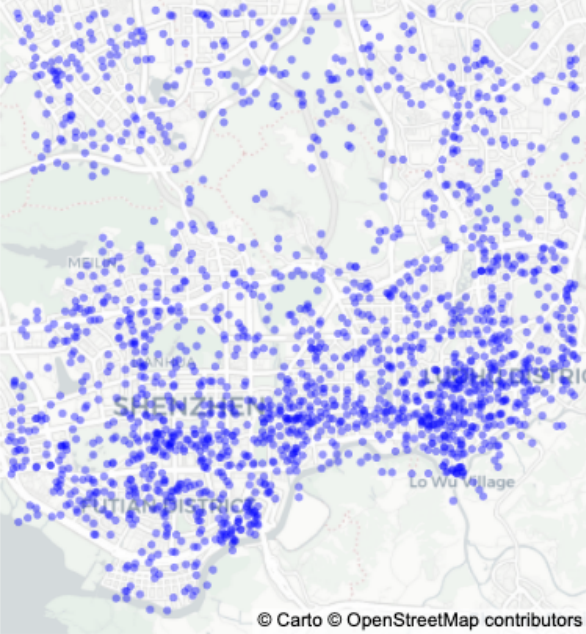}
    \captionsetup{justification=centering}
    \caption{$\alpha=0.4$}
\end{subfigure}
\hfill
\begin{subfigure}[t]{0.16\textwidth}
    \centering
    \includegraphics[width=0.98\textwidth]{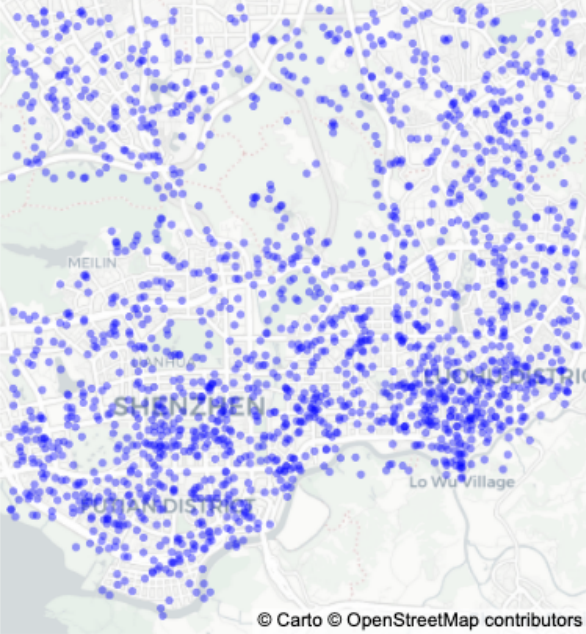}
    \captionsetup{justification=centering}
    \caption{$\alpha=0.6$}
\end{subfigure}
\hfill
\begin{subfigure}[t]{0.16\textwidth}
    \centering
    \includegraphics[width=0.98\textwidth]{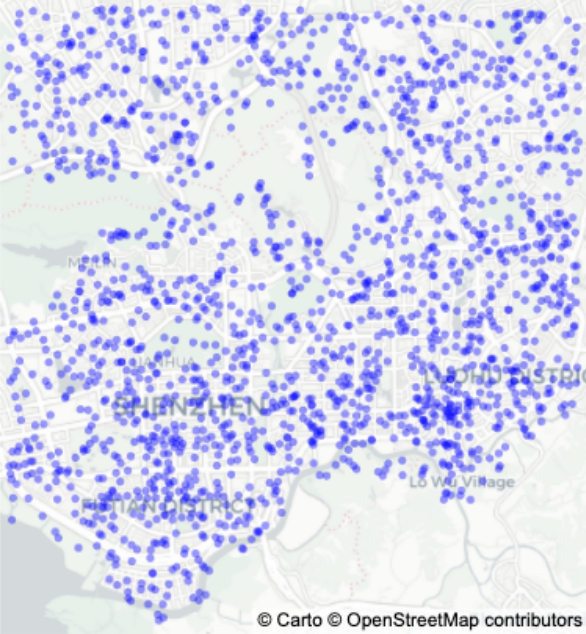}
    \captionsetup{justification=centering}
    \caption{$\alpha=0.8$}
\end{subfigure}
\hfill
\begin{subfigure}[t]{0.16\textwidth}
    \centering
    \includegraphics[width=0.98\textwidth]{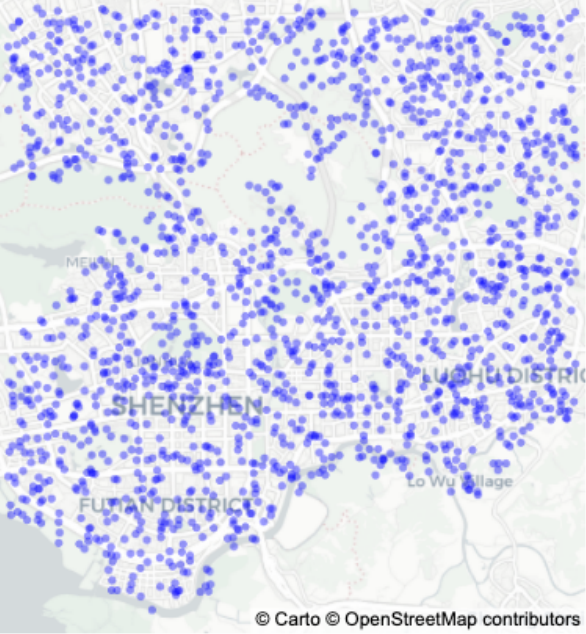}
    \captionsetup{justification=centering}
    \caption{$\alpha=1$}
\end{subfigure}
\caption{Initial vehicle distribution $\mu_0$ against parameter $\alpha$.}
\label{fig:vehicles_initializations}
\end{figure*}

To offer flexible initialization strategies during the training and to examine the sensitivity of vehicle rebalancing with respect to the initial vehicle distribution during the evaluation, we initialize $\mu_0$ as a linear combination of a uniform distribution $U$ and the demand distribution at time $t=0$, i.e., 
\begin{align} \label{eq:vehicles_initialization}
    \mu_0 = \alpha U + (1-\alpha)\odelta_0,  
\end{align}
where $\alpha \in [0, 1]$ represents the fraction of uniformly initialized vehicles (see \Cref{fig:vehicles_initializations}). 
For example, this variation lets us compare vehicle initialization strategies like fully demand-driven ($\alpha=0$) versus uniform ($\alpha=1$), where drivers begin their shifts from random locations.
As later explained in \Cref{sec:learning-protocol-setup}, these two extreme cases are used in model training. 

The vehicle and mean-field movements follow the transition functions specified in \Cref{sec:transition}. To reflect the real-world practice, we set no noise to the Gaussian associated with matched vehicles $\epsilontm$, which implies riders are dropped off exactly at their destinations. On the other hand, a truncated Gaussian noise with a standard deviation of $0.0175$ is assigned to $\epsilontr$ and $\epsilontc$ that correspond to the repositioning and cruising vehicles, respectively. It then yields vehicles ending within a 500-meter radius of their desired location with high probability. 

\subsubsection{Matching Module}\label{sec:matching-module}
\begin{figure}[htb!]
    \centering
    \includegraphics[width=0.6\textwidth, trim=10cm 6cm 10cm 7.2cm, clip]{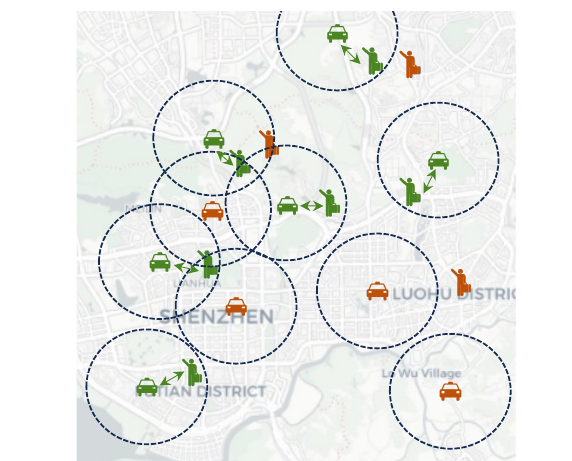}
    \caption{Schematic representation of the matching module. Dotted circles represent the vehicles' matching radius. In green are vehicle-rider matching pairs, and in red are those that failed to match.}
    \label{fig:matching_module}
\end{figure}

To represent the underlying matching process $M$, we integrate a matching module, depicted in \Cref{fig:matching_module}, into the simulator that performs bipartite matching between waiting riders and available vehicles at every minute. During each 20-minute interval, the ride requests are generated from the demand measure $\deltat$ uniformly over the interval with a 5-minute maximum waiting time. The matching problem is formulated as an assignment problem~\citep{burkard2012assignment,ramshaw2012minimum} with the objective of minimizing the total Euclidean pickup distance between available vehicles and riders. A maximum pickup distance of around 850 meters is also introduced as a constraint to prevent long pickup times. \looseness=-1

\subsubsection{Accessibility Constraint Specification}
\begin{figure*}[htb!]
\centering
\begin{subfigure}[t]{0.19\textwidth}
    \centering
    \includegraphics[width=0.98\textwidth]{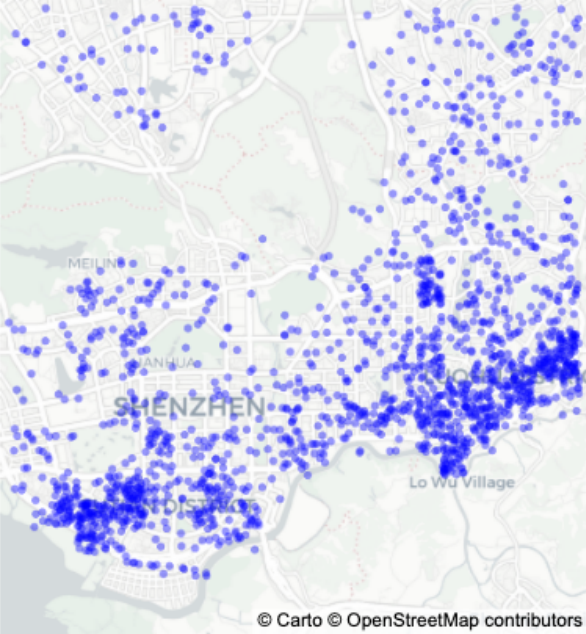}
    \captionsetup{justification=centering}
    \caption{$p=0$}
\end{subfigure}
\hfill
\begin{subfigure}[t]{0.19\textwidth}
    \centering
    \includegraphics[width=0.98\textwidth]{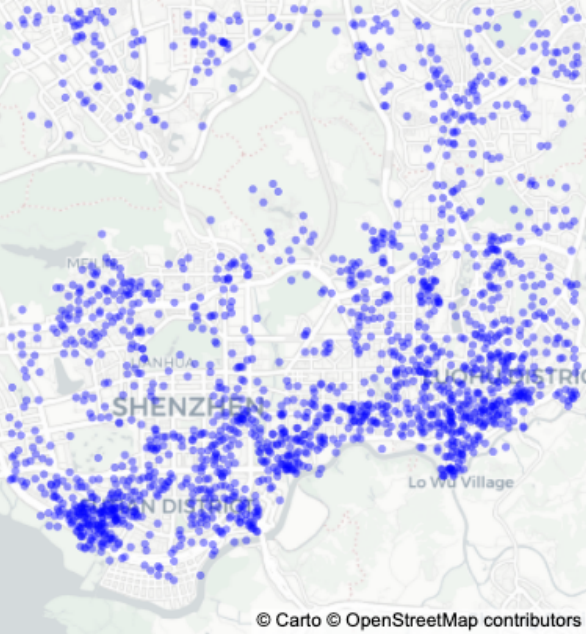}
    \captionsetup{justification=centering}
    \caption{$p=0.2$}
\end{subfigure}
\hfill
\begin{subfigure}[t]{0.19\textwidth}
    \centering
    \includegraphics[width=0.98\textwidth]{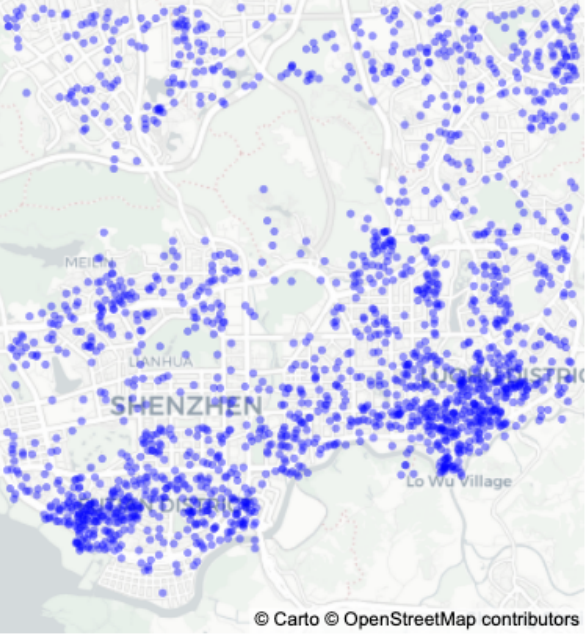}
    \captionsetup{justification=centering}
    \caption{$p=0.4$}
\end{subfigure}
\hfill
\begin{subfigure}[t]{0.19\textwidth}
    \centering
    \includegraphics[width=0.98\textwidth]{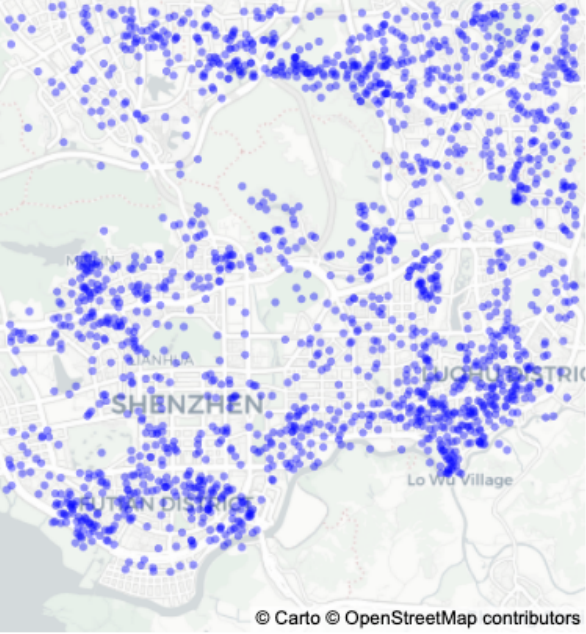}
    \captionsetup{justification=centering}
    \caption{$p=0.6$}
\end{subfigure}
\hfill
\begin{subfigure}[t]{0.19\textwidth}
    \centering
    \includegraphics[width=0.98\textwidth]{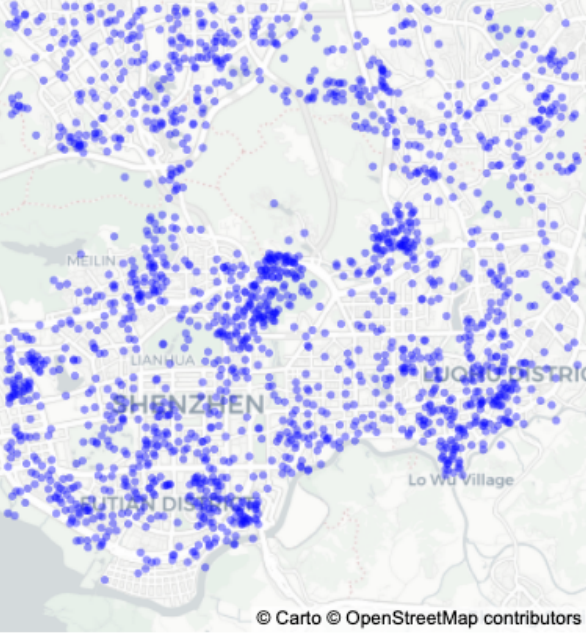}
    \captionsetup{justification=centering}
    \caption{$p=0.8$}
\end{subfigure}

\caption{Vehicle distributions $\mu_T$ at the end of interval 21:40-22:00 at different accessibility constraints.}
\label{fig:vehicles_distributions}
\end{figure*}

To specify the accessibility constraint, we first compute the maximum of $\varepsilon$-smoothed weighted differential entropy given the weight function $w(\cdot)$:
\begin{align}
    \hw_{\max} = \max_{\mu\in\pp(\s)}-\int_\sis w(s)\log(\mu(s) + \varepsilon)ds.
\end{align}
Then, we specify different accessibility lower bounds $C=p\hw_{\max}$ for some factor $p \in [0,1]$. Accordingly, when $p=0$, no accessibility constraint is imposed on the vehicle rebalancing operations, and thus, the platform can freely match the vehicle supply with the ride demand to maximize the profit. In contrast, when $p=1$, the platform is forced to achieve the maximum entropy of vehicle distribution through rebalancing, leading to equally accessible service at every location. 
An example of vehicle distribution at different accessibility constraints is illustrated in \Cref{fig:vehicles_distributions}.

\subsection{Model Training}\label{sec:training}
The training procedure consists of two parts: (i) learning state transition, i.e., matching process, and (ii) optimizing vehicle rebalancing policy.  

\subsubsection{Matching Process Classifier} \label{sec:matching-process-classifier}
When MFC is adopted, the former is not needed because the matching process $M$ is directly approximated by the optimal transport $\hM$. Thanks to the discrete representation of (probability) measures, the original optimal transport is turned into a minimum cost flow problem~\citep{ahuja1993network,ford2015flows}, as detailed in \Cref{apx:discrete-ot}. 
For MFRL, the matching process classifier $\tM$ is learned from the more realistic matching outcomes obtained through the interactions with the matching module described in \Cref{sec:matching-module}, as explained in \Cref{sec:learning_matching_process}. 
We parametrize the matching process binary classifier $\tM$ via a fully connected neural network and train it to minimize cross-entropy loss with AdamW optimizer \citep{loshchilov2017decoupled} on the matching outcomes stored in the experience replay buffer.
We leverage insights from \citet{jusup2024safe}, which show that increasing the number of data-collecting vehicles significantly accelerates learning, thereby reducing computational costs.
Thus, we collect a data sample per zone at each time step instead of only using observations from a representative vehicle.
This aligns with real practice, where the ride-sourcing platform can access all matching outcomes over its fleet. The hyperparameters used in the training of the matching process are summarized in \Cref{tab:classifier_hyperparameters}. 

\begin{table*}[t!]
\centering
\caption{Hyperparameters used during the MFRL training of the matching process classifier.}
\label{tab:classifier_hyperparameters}
\small
\begin{tabular}{l|r|l}
\toprule
Hyperparameter & Value & Description \\
\midrule
\# of hidden layers & 5 &   \\
\# of neurons & 2056, 1024, 256, 64, 8 & Number of neurons per hidden layer \\
hidden activations & Leaky-ReLU & \\
output activation & Sigmoid & \\
$\alpha$ & $10^{-4}$ & Learning rate \\
$w$ & $5\cdot 10^{-4}$ & Weight decay \\
weights initialization & Kaiming uniform & \\
bias initialization & 0 & \\
$K$ & 1,000 & Number of epochs \\
replay buffer size & 10 & Stores the observations for the last 10 episodes \\
batch size & 8, 16, 32 & \makecell[l]{Progressively increasing with the number of samples \\ in replay buffer} \\
train-validation split & 90\%-10\% & We use a validation set for early stopping \\
early stopping patience threshold & 30 & Terminate training if the threshold has been reached \\
minimum improvement & 1\% & \makecell[l]{If not improved after an epoch, increase patience by 1, \\ otherwise reset} \\
\bottomrule
\end{tabular}
\end{table*}

\subsubsection{Policy Profile} \label{sec:policy-profile}

As for the learning of optimal policy, we first reformulate \Cref{eq:log-barrier} into a lifted \cmfmdp proposed in \citet{jusup2024safe}. Instead of introducing $T$ neural networks (one per time step) to represent the time-dependent policy profile $\bpin$, we define a single policy network that also takes the time step as input in addition to the state and mean-field variables. 
We then use the mean-field variant of back-propagation-through-time (MF-BPTT) introduced in \citet{jusup2024safe} for the weight updates. 
The policy network is initialized using Kaiming uniform initialization~\citep{he2015delving} and trained for 1,000 epochs with early stopping when the policy does not show sufficient improvement. 
To stabilize the training process, we use L2-norm gradient clipping to prevent gradient explosions and implement the post-activation batch normalization \citep{ioffe2015batch} to prevent vanishing gradients. 
We further enforce exploration by adding a truncated Gaussian with standard deviation following exponential decay to the policy outputs. 
Finally, AdamW is used as the optimizer, along with hyperparameters summarized in \Cref{tab:policy_hyperparameters}.

\begin{table*}[t!]
\centering
\caption{Hyperparameters for training neural network policy.}
\label{tab:policy_hyperparameters}
\small
\begin{tabular}{l|r|l}
\toprule
Hyperparameter & Value & Description \\
\midrule
\# of hidden layers & 4 &   \\
\# of neurons & 512, 256, 256, 64 & Number of neurons per hidden layer \\
hidden activations & Leaky-ReLU & \\
output activation & Tanh & \\
$\alpha$ & $5 \cdot 10^{-4}$ & Learning rate \\
$w$ & $5\cdot 10^{-4}$ & Weight decay \\
weights initialization & Kaiming uniform & \\
bias initialization & 0 & \\
exploration decay & 0.02 & \\
$K$ & 1,000 & Number of epochs \\
early stopping patience threshold & 1 & Terminate training if the threshold has been reached \\
minimum improvement & 5\% & \makecell[l]{If not improved after 100 epochs, increase patience by 1, \\ otherwise reset} \\
\bottomrule
\end{tabular}
\end{table*}

\subsubsection{Learning Protocol Setup} \label{sec:learning-protocol-setup}
To better distinguish different objectives of vehicle rebalancing, we train two sets of policies using the two extreme cases of initial vehicle distribution $\mu_0$: (i) profit-driven ($p=0$) with uniform distribution ($\alpha=1$), and (ii) accessibility-driven ($p>0$) with demand distribution ($\alpha=0$). 
In other words, the model is trained in the scenarios where the most rebalancing efforts are needed. Otherwise, the model would immediately get a high reward without sufficient exploration. In this way, we expect the resulting policies to properly handle all possible situations and display robustness in various scenarios, which will be further investigated in \Cref{sec:robustness-test}. 
All training scenarios use a fixed log-barrier hyperparameter $\lambda=1$, as justified in \Cref{apx:impact-of-lambda}.
The rest of the learning protocol hyperparameters are summarized in \Cref{tab:learning-protocol-hyperparameters}.

\begin{table*}[t!]
\centering
\caption{Hyperparameters used as inputs to the MFRL learning protocol.}
\label{tab:learning-protocol-hyperparameters}
    \small
    \begin{tabular}{l|r|l}
    \toprule
    Hyperparameter  & Value & Description \\
    \midrule
    $N$ & 10 & Number of episodes \\
    $T$ & 18 & Number of time steps \\
    $M \times M$ & $25 \times 25$ & Grid size  \\
    $\sigma^{R}, \sigma^{M}, \sigma^{C}$ & 0.0175, 0, 0.0175  & \makecell[l]{Standard deviations of the idiosyncratic noise of repositioning, \\ matched and cruising vehicles}  \\
    $\lambda$ & 1 & Log-barrier hyperparameter \\
    $\varepsilon$ & $10^{-10}$ & $\varepsilon$-smoothing of the differential entropy \\
    \# of representative vehicles & 471 & One per operational zone \\
    \bottomrule
    \end{tabular}
\end{table*}

\subsection{Evaluation Setup} \label{sec:benchmarks_and_evals}
We focus on measuring business impact, benchmarking against relevant policies, and conducting extensive ablation studies.
The training phase utilizes the population mean-field distribution $\mu$ to derive policies $\bpisot$ and $\bpis_N$. During evaluation, these policies are used to control the empirical mean-field distribution $\mu^L$ of a finite fleet of $L$ vehicles. The resulting approximation error of the objective function in \Cref{eq:mfc-objective} between using $\mu$ and $\mu^L$ is theoretically bounded as shown in \citet{Gu2021Mean-FieldAnalysis}, and empirically investigated in \cite{jusup2024safe}. Specifically, \mfcp{p} and \mfrlp{p}, together with benchmark policies, are evaluated over 10 simulation runs controlling $L=18{,}000$ vehicles. The fleet size $L$ is inferred as a 20-minute fleet-size average from the dataset and rounded to the nearest thousand.

\subsubsection{Benchmarks} \label{sec:benchmarks}

We compare MFC and MFRL-based policies subject to different accessibility constraints defined by a fraction $p$ of the maximum entropy $\hw_{\max}$, denoted as \mfcp{p} and \mfrlp{p}, to three benchmark policies:
\begin{itemize}
    \item \textbf{No rebalancing:} All vehicles remain cruising locally in the same zone until getting matched with new riders. It is used to demonstrate the necessity of vehicle rebalancing. 

    \item \textbf{Static rebalancing (LP-static):}  A static rebalancing strategy is a linear program following \citet{pavone2012robotic}, which aims to maintain the same vehicle supply in each zone by rebalancing vehicles to compensate for the imbalance between inflows and outflows of rider trips. This policy implicitly ensures the accessibility constraint as the desired vehicle supply in each zone. More details about this benchmark are discussed in \Cref{apx:static-rebalancer}. \looseness=-1

    \item \textbf{Dynamic rebalancing (LP-dynamic):} As an extension to the static rebalancing policy, the desired vehicle supply is specified to be time-varying, and the rebalancing strategy is solved considering dynamic demand patterns. For further details, see \Cref{apx:dynamic-rebalancer}.
    
\end{itemize}

\subsubsection{Metrics} \label{sec:metrics}
For the performance evaluation and comparison, we define a set of metrics that covers the system level and both sides of the market, allowing us to generate insights from different perspectives:
\begin{itemize}
    \item \textbf{System level}
    \begin{itemize}
        \item Service accessibility: the fraction of zones with at least one available vehicle 
        \item Service fulfillment: the fraction of zones with at least 90\% of ride requests being satisfied
    \end{itemize}
    \item \textbf{Supply side}
    \begin{itemize}
        \item Utilization rate: the fraction of matched vehicles
        \item Rebalancing rate: the fraction of vehicles under repositioning
    \end{itemize}
    \item \textbf{Demand side}
    \begin{itemize}
        \item Pickup distance: the average distance between matched vehicles and riders
        \item Service rate: the fraction of satisfied requests 
    \end{itemize}
\end{itemize}

\subsection{Results} \label{sec:results}

All results discussed in this section are the average over 10 evaluation runs, and the standard deviations are only reported when necessary. In what follows, we will first compare the computation time of proposed mean-field methods and benchmarks (\Cref{sec:computation-time}), then discuss the tradeoffs between accessibility and other service quality metrics (\Cref{sec:accessibility-tradeoff}). Finally, we will present a series of tests on model robustness (\Cref{sec:robustness-test}).

\subsubsection{Computational Efficiency}\label{sec:computation-time}

\begin{figure*}[htb!]
\centering
\includegraphics[width=0.5\textwidth]{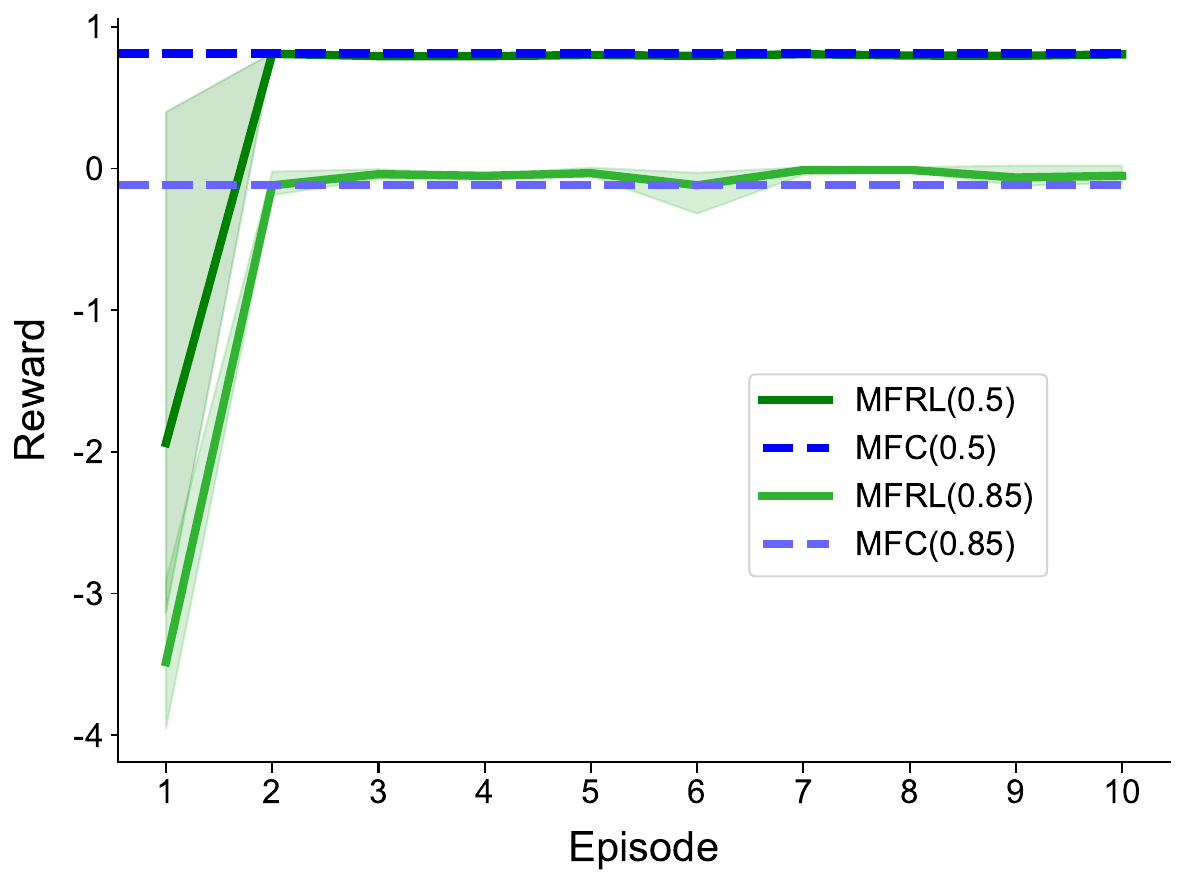}
\captionsetup{justification=justified}
\caption{Learning curves of MFRL subject to different accessibility constraints. The average rewards (solid line) and standard deviations (shaded area) over 4 training runs with randomly initialized neural network weights are plotted, along with the final MFC rewards (dashed line).
}
\label{fig:learning_curves}
\end{figure*}

\Cref{fig:learning_curves} plots the learning curves of \mfrlp{0.5} and \mfrlp{0.85} along with the final rewards of the corresponding MFC models, \mfcp{0.5} and \mfcp{0.85}, respectively. Note that MFC only needs one episode of training because the matching process is fixed, and it serves as a benchmark of MFRL. Both MFRL policies converge within only two episodes and achieve a similar reward to MFC policies. The efficient learning is largely attributed to the access to observations in all zones, as described in \Cref{sec:matching-process-classifier}.

\Cref{tbl:compute_time} reports the training times of MFC and MFRL and the inference times of all tested rebalancing policies. Specifically, we evaluate MFC and MFRL on both GPU and CPU, while the benchmark policies are only implemented on CPU. 
Overall, the training process of both MFC and MFRL is efficient, finishing within 40 minutes and 26 minutes, respectively. 
It is worth noting that the major bottleneck of MFC training is solving the minimum cost flow problem, which is currently executed on the CPU, whereas training the policy on the GPU takes just over a minute in total.
Hence, an end-to-end GPU implementation of MFC shall significantly reduce the computation time and is expected to outperform MFRL, as the latter must train both policy and state transition functions over multiple iterations. Specifically, learning the matching process amounts to 34\% of the training time for MFRL. 
In \Cref{tbl:compute_time}, we report averages and standard deviations across multiple values of parameter $p$ (see \Cref{sec:accessibility-constraint}), albeit it takes slightly longer to optimize the policy as the accessibility constraint becomes stricter, i.e., $p$ increases. 

In terms of the inference time for 18,000 vehicles, i.e., time to generate rebalancing policy, both MFC and MFRL are far more efficient than the benchmarks, regardless of whether they are run on CPU or GPU. Specifically, on a GPU, they run within a fraction of a second instead of the tens of minutes required by benchmarks. This demonstrates the great potential of MFC and MFRL-based vehicle rebalancing in real practice. 

\begin{table*}[htb!]
\centering
\caption{Computation time for training and inference.}
\label{tbl:compute_time}
\begin{tabular}{l|cc|cc|c|c}
\toprule
Model &  \multicolumn{2}{c|}{MFC} & \multicolumn{2}{c|}{MFRL} & LP-static & LP-dynamic  \\
\midrule
Hardware & GPU \& CPU & CPU & GPU & CPU & CPU & CPU \\
Training time (min) & 39.6 $\pm$ 2.7 & N/A & 25.4 $\pm$ 1.5 & N/A & \\
Inference time (min) & <0.001 & 0.25 & <0.001 & 0.25  & 18.9 & 12.3 \\
\bottomrule
\end{tabular}
\end{table*}

\subsubsection{Tradeoff Between Accessibility and Other Metrics}\label{sec:accessibility-tradeoff}

\begin{figure*}[h!]
\centering
\begin{subfigure}{0.49\textwidth}
    \centering
    \includegraphics[width=0.98\textwidth]{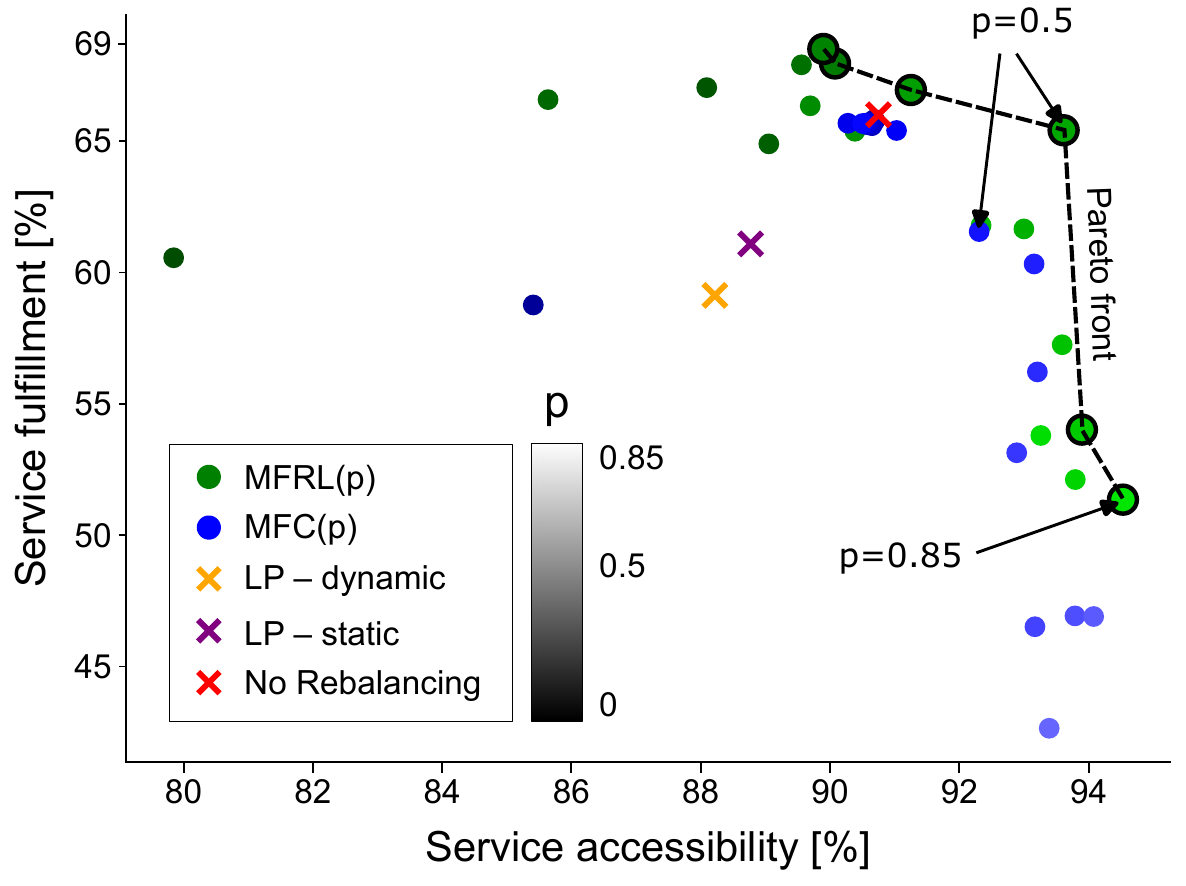}
    \captionsetup{justification=centering}
    \caption{vs. service fulfillment}
    \label{fig:accessibility-fullfillment}
\end{subfigure}
\hfill
\begin{subfigure}{0.49\textwidth}
    \centering
    \includegraphics[width=0.98\textwidth]{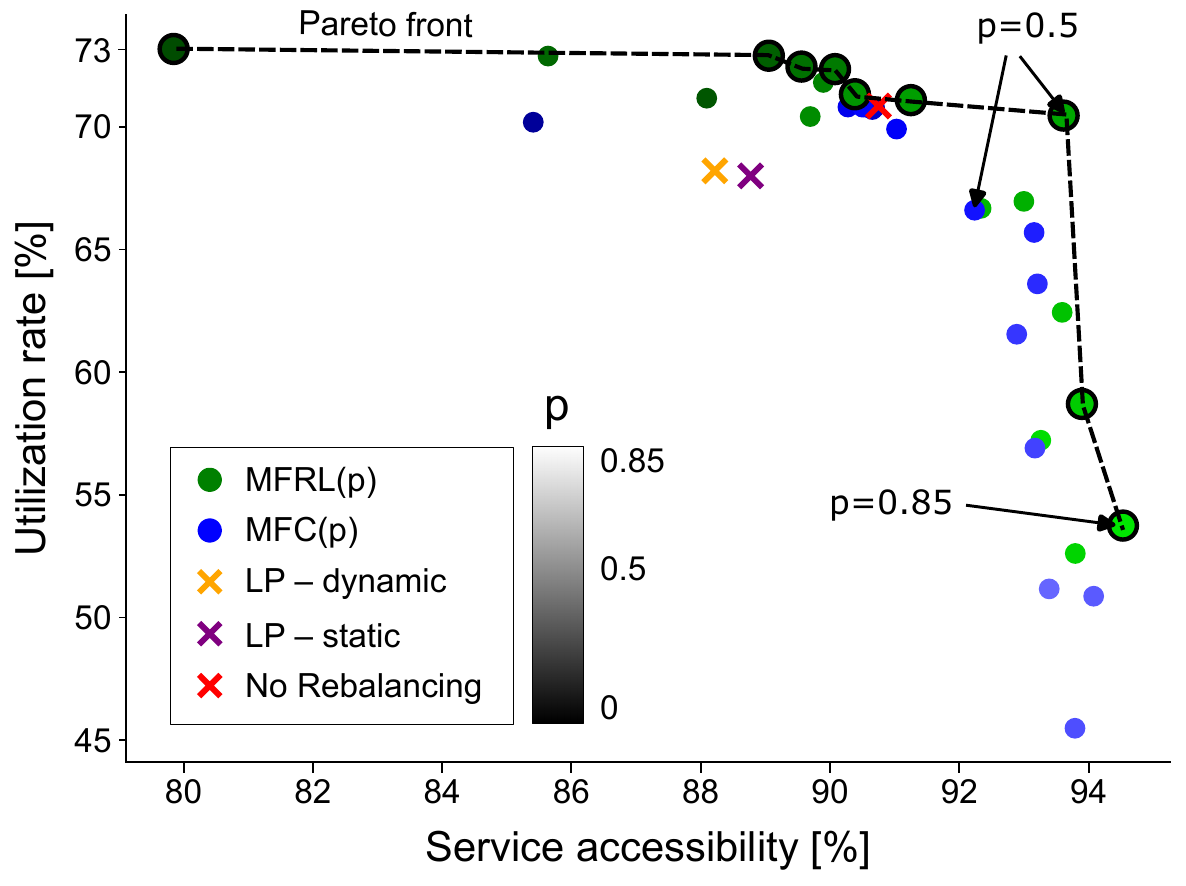}
    \captionsetup{justification=centering}
    \caption{vs. vehicle utilization rate}
    \label{fig:accessibility-matched}
\end{subfigure}
\par\vspace{0.3cm} 
\begin{subfigure}{0.49\textwidth}
    \centering
    \includegraphics[width=0.98\textwidth]{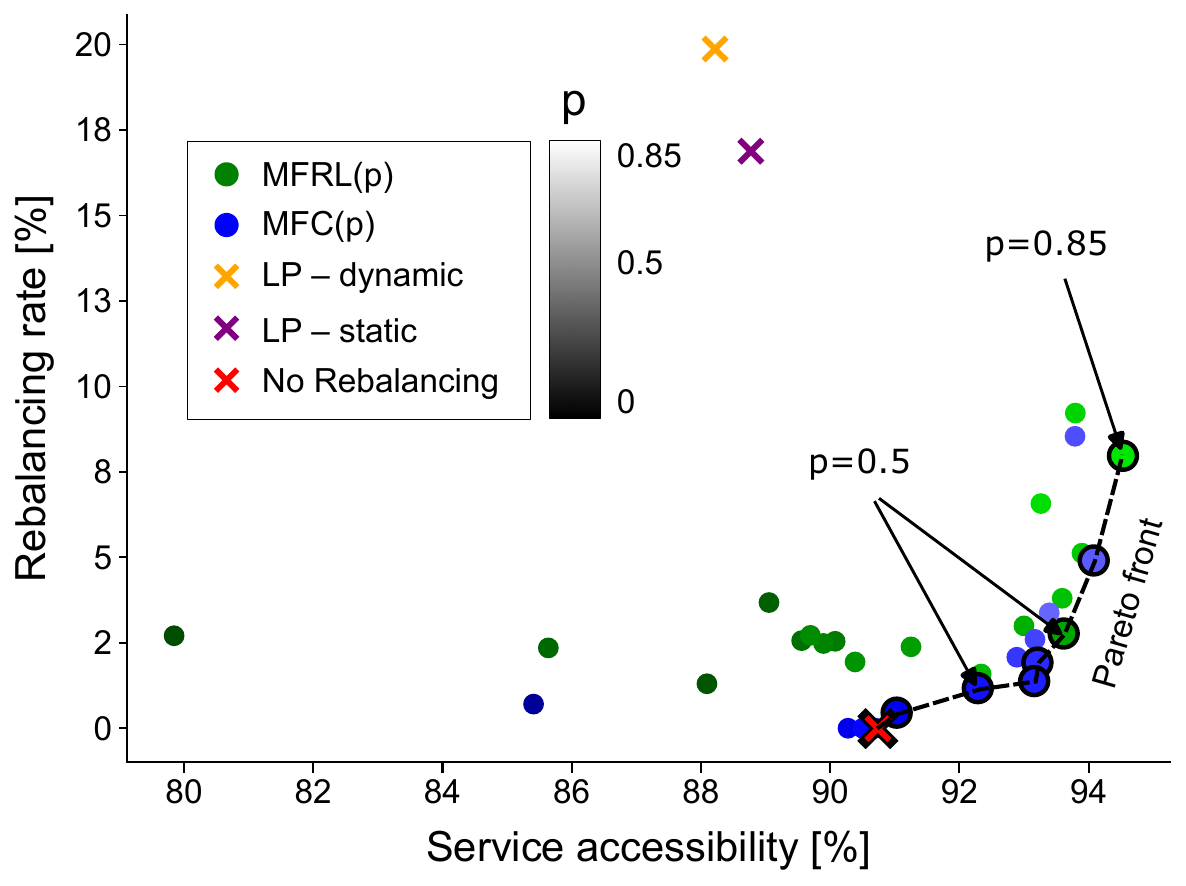}
    \captionsetup{justification=centering}
    \caption{vs. vehicle rebalancing rate}
    \label{fig:accessibility-repositioned}
\end{subfigure}
\hfill
\begin{subfigure}{0.49\textwidth}
    \centering
    \includegraphics[width=0.98\textwidth]{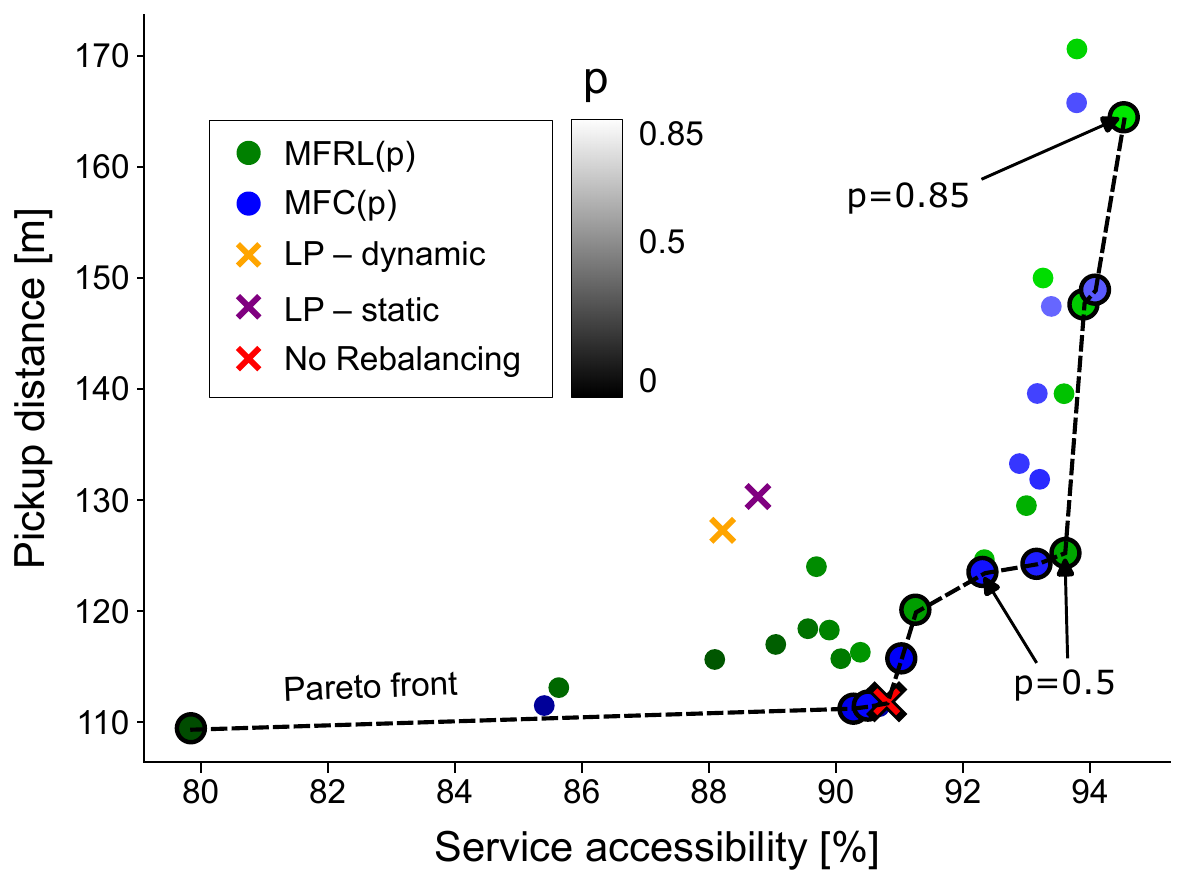}
    \captionsetup{justification=centering}
    \caption{vs. pickup distance}
    \label{fig:accessibility-distance}
\end{subfigure}
\par\vspace{0.3cm} 
\begin{subfigure}{0.49\textwidth}
    \centering
    \includegraphics[width=0.98\textwidth]{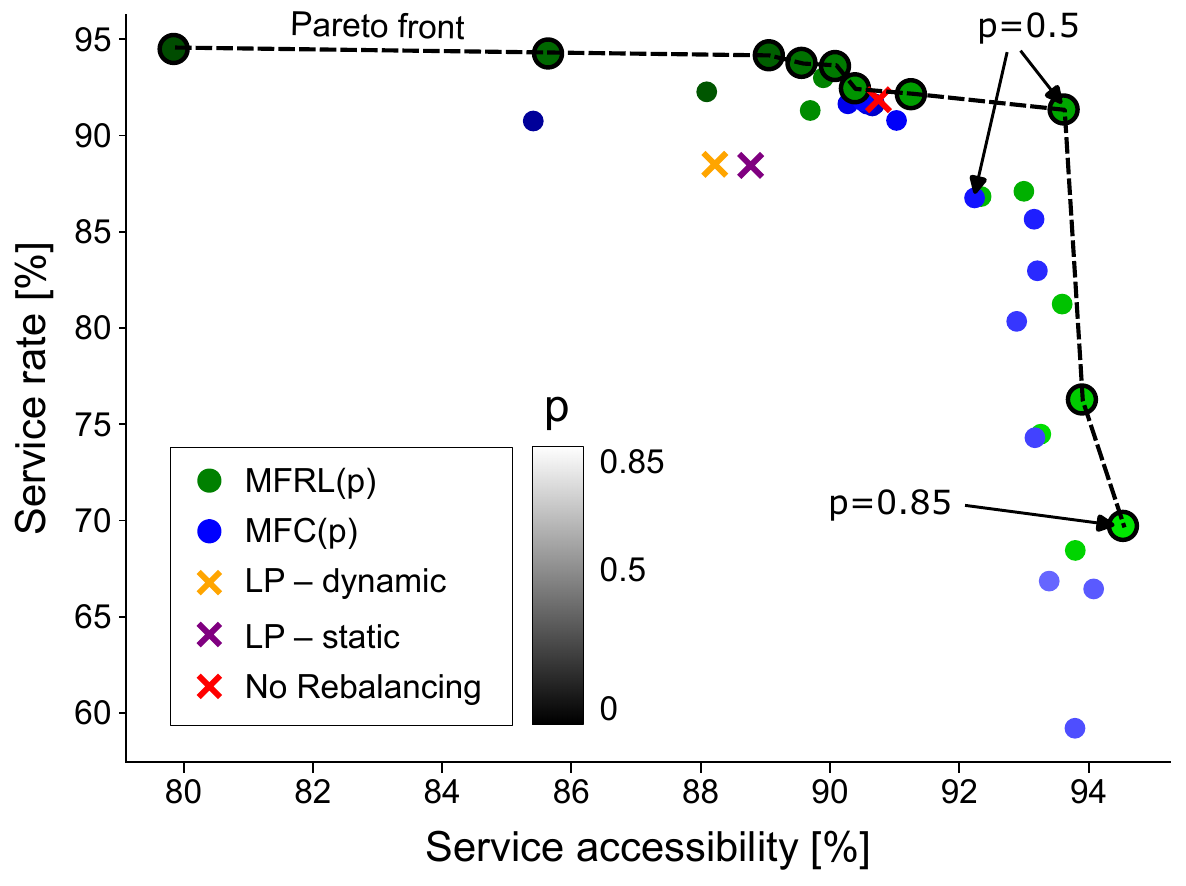}
    \captionsetup{justification=centering}
    \caption{vs. service rate}
    \label{fig:accessibility-satisfied}
\end{subfigure}
\caption{Comparison of service accessibility against other performance metrics. We indicate Pareto fronts (dashed lines) and \mfrlp{0.5}, \mfrlp{0.85}, and \mfcp{0.5} for further ablations.} 
\label{fig:performance-original_demand}
\end{figure*}

Recall that the main objective of this study is to achieve a balance between service accessibility and other performance metrics. 
Hence, we train and evaluate the model with different levels of accessibility constraints by varying the value of $p\in [0,0.85]$. \Cref{fig:performance-original_demand} illustrates the performance of all tested policies against the fraction of accessible zones and other metrics, along with the Pareto fronts. 

Our first observation from \Cref{fig:performance-original_demand} is that no rebalancing already achieves a reasonably good balance between service accessibility and other performance metrics (e.g., around 65\% service fulfillment and 70\% utilization rate). This is largely due to the balanced demand pattern in the tested service region. In contrast, neither static nor dynamic LP approaches can beat no rebalancing, which implies they fail to capture the supply-demand dynamics in the market. A significant downside of no rebalancing is that it limits decision-making by not offering any control over service outcomes. \looseness=-1

\Cref{fig:performance-original_demand} also clearly illustrates the trade-off between service accessibility and other performance metrics. As the accessibility constraint becomes stricter, i.e., $p$ increases, the fraction of accessible zones increases, while all other metrics deteriorate. Nevertheless, when MFRL and MFC-based rebalancing policies are implemented, the performance only drops significantly when very high accessibility is required. 
For instance, by increasing $p$ from zero to 0.5, we lose less than 5\% in the vehicle utilization rate under the best case. These results indicate that ride-sourcing platforms can easily achieve desirable service accessibility or, equivalently, ensure certain mobility equity without greatly sacrificing their service efficiency and profitability. 

Compared to MFC, MFRL remains on the majority of Pareto fronts for service fulfillment, vehicle utilization, and service rate, though it tends to produce more vehicle rebalancing orders and longer pickup distances. Therefore, MFRL could be a preferred choice if the operational budget is not tightly limited. Particularly, it not only achieves better service performances than the LP-based policies but its low inference time enables real-time responses to market changes and effective rebalancing decisions (see \Cref{tbl:compute_time}). 
On the other hand, MFC could also be an ideal option when less training time is available, as its performance gap from MFRL is not substantial.

\subsubsection{Robustness Analyses}\label{sec:robustness-test}

In this section, we present the main results of a series of robustness tests using \mfrlp{0.5}, \mfrlp{0.85}, and \mfcp{0.5} models. We select \mfrlp{0.5} and \mfrlp{0.85} because they stand on the Pareto front for every metric in \Cref{fig:performance-original_demand}, while \mfrlp{0.85} is also subject to the highest accessibility constraint. Besides, \mfcp{0.5} is tested, as it frequently appears on the Pareto front and is considered a good benchmark for \mfrlp{0.5} with the same accessibility constraint. \looseness=-1

\begin{figure*}[htb!]
\centering
\includegraphics[width=0.8\textwidth]{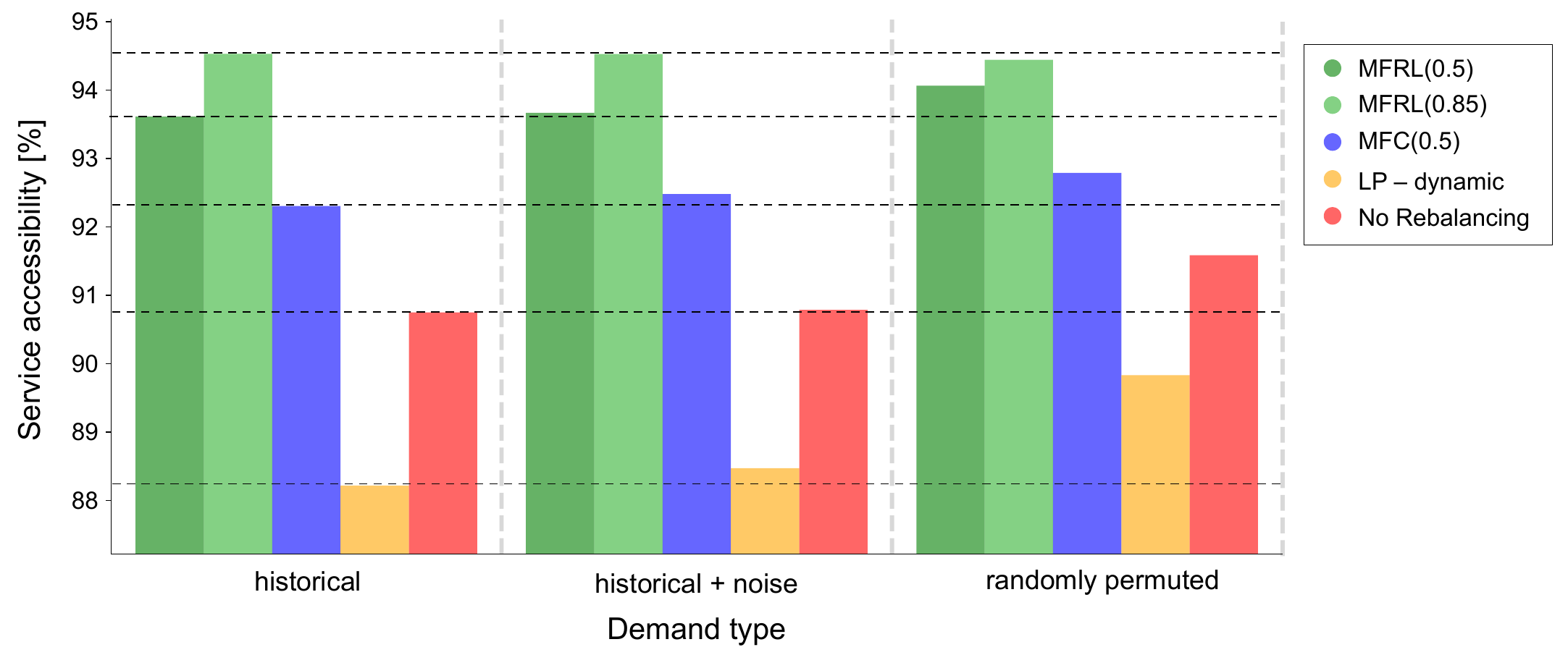}
\captionsetup{justification=justified}
\caption{Fraction of accessible zones under different tested demand patterns.}
\label{fig:robustness_demand}
\end{figure*}

\paragraph{Robustness to unforeseen demand patterns}
To evaluate the robustness toward unforeseen demand patterns, we test models trained on the historical demand in the other three scenarios described in \Cref{sec:demand_pattern}. 
\Cref{fig:robustness_demand} reports the accessibility achieved in the first three demand scenarios. It can be seen that both MFRL and MFC policies are able to maintain accessibility under the scenarios described in \Cref{fig:demand_patterns}, of slight spatial demand perturbations and significantly shifted temporal demand patterns. Overall, the accessibility under Gaussian noise is particularly stable. At the same time, the randomly permuted demand pattern even improves service accessibility, which also increases under no rebalancing, indicating that the spatial demand structure might be more straightforward. Accordingly, the accessibility under \mfrlp{0.5} and \mfcp{0.5} slightly increases as well. Although almost negligible, the accessibility under \mfrlp{0.85} decreases when the demand is randomly permuted. As \mfrlp{0.85} repositions vehicles more intensively, slight performance deviations are expected to maintain high accessibility. \looseness=-1

We now proceed to analyze the robustness of the models towards occasional demand shocks. This is motivated by temporary demand outbursts from events such as concerts, sports, and other public gatherings. These events usually occur in areas with low demand outside the city center (see \Cref{fig:requests_shock}), where few vehicles are present on a typical day. The hypothetical demand shock is designed to fit these criteria, and it is used to explore whether the rebalancing policies trained on nominal demand patterns still work in these scenarios. 

\begin{figure*}[h!]
\centering
\begin{subfigure}[t]{1\textwidth}
    \centering
    \includegraphics[width=0.98\textwidth]{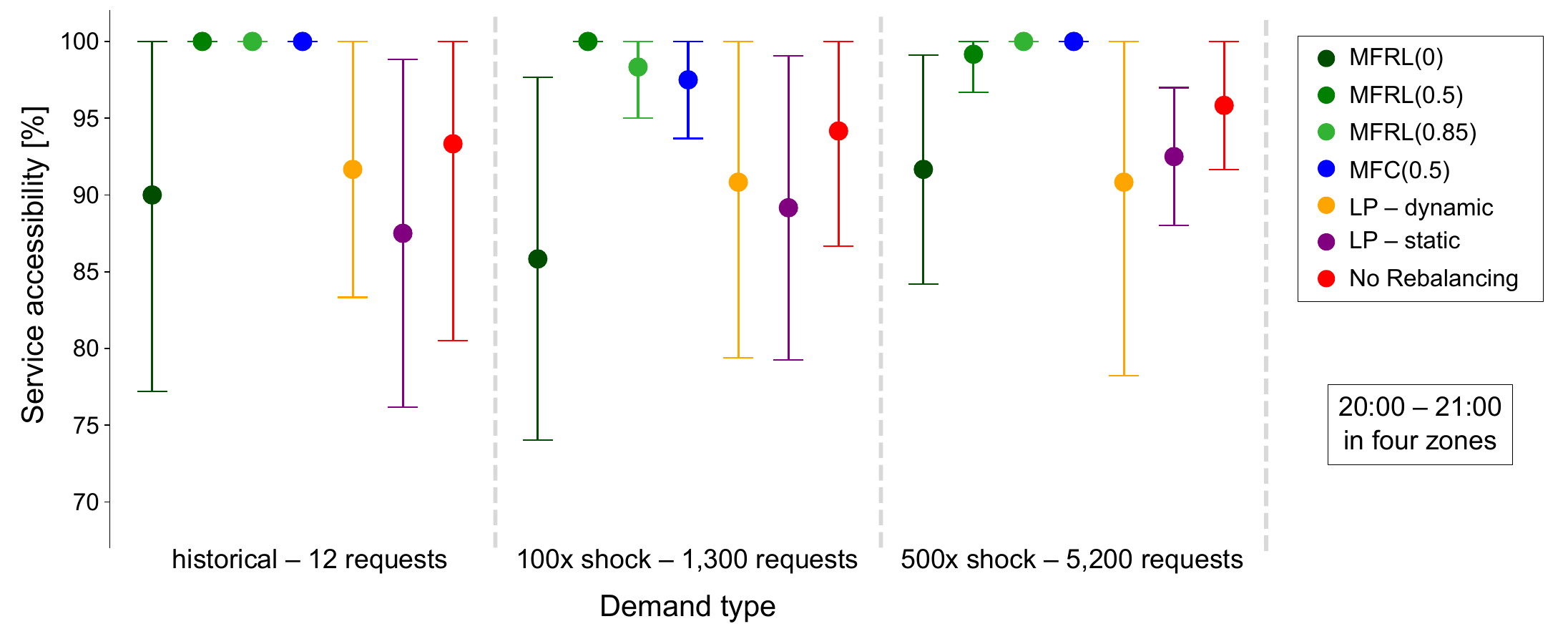}
    \captionsetup{justification=centering}
    \caption{Fraction of accessible zones}\label{fig:performance_shock_demand_accessibility}
\end{subfigure}
\par\vspace{0.5cm} 
\begin{subfigure}[t]{1\textwidth}
    \centering
    \includegraphics[width=0.98\textwidth]{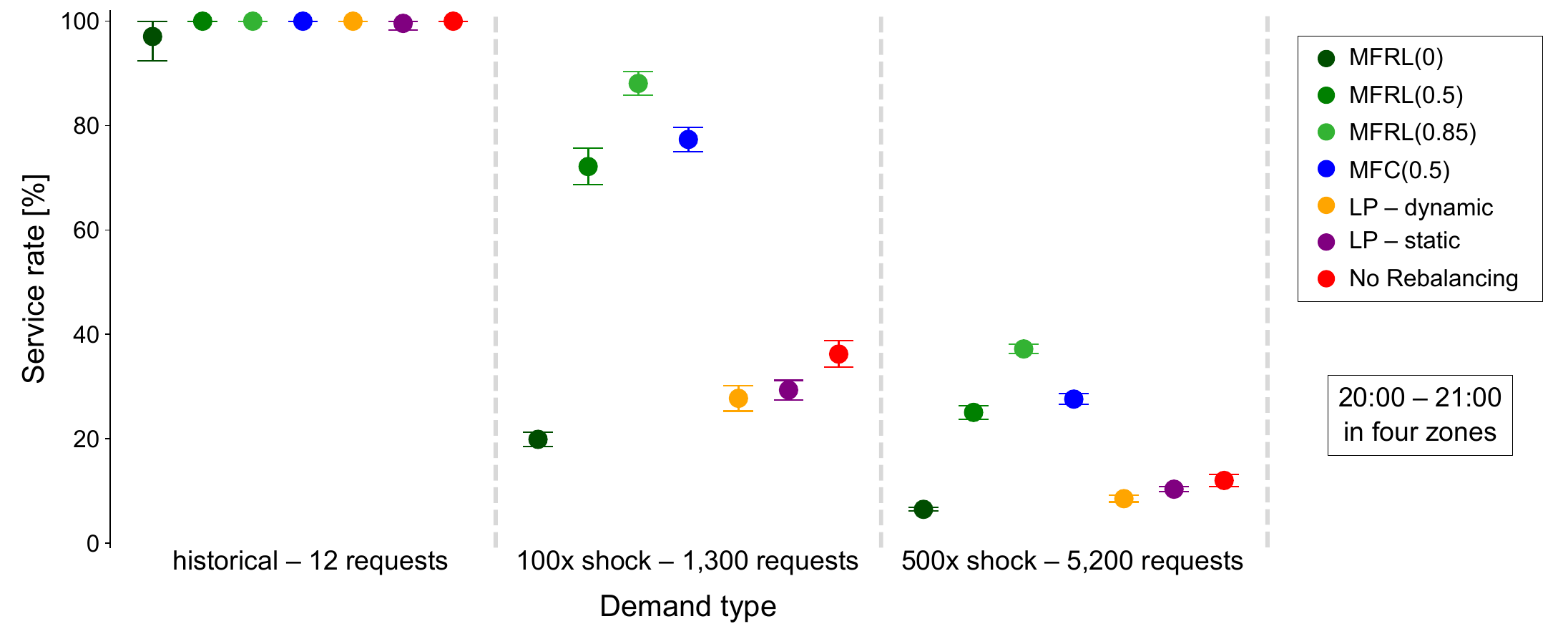}
    \captionsetup{justification=centering}
    \caption{Service rate}\label{fig:performance_shock_demand_service_rate}
\end{subfigure}
\par\vspace{0.5cm} 
\begin{subfigure}[t]{1\textwidth}
    \centering
    \includegraphics[width=0.98\textwidth]{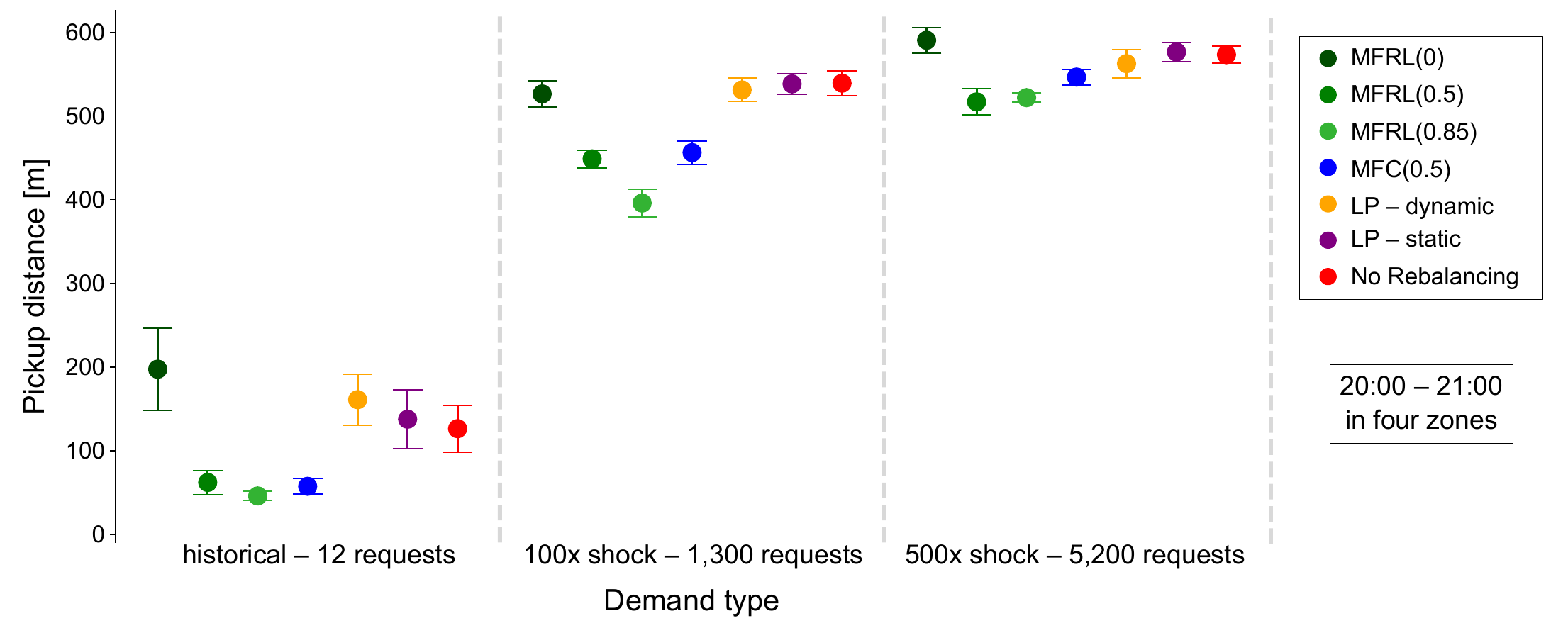}
    \captionsetup{justification=centering}
    \caption{Pickup distance}\label{fig:performance_shock_demand_pickup_distance}
\end{subfigure}
\caption{Comparison of service performances under demand shocks.}
\label{fig:performance-shock_demand}
\end{figure*}

\Cref{fig:performance-shock_demand} illustrates the accessibility, service rate, and pickup distance within the zones of interest under both historical demand and demand shocks. In the historical data, very few requests are observed in the studied zone and time period, while we manually scale it up to 1,300 and 5,200 requests in the hypothetical demand shocks. 
As shown in \Cref{fig:performance_shock_demand_accessibility}, under the historical demand, all MFC and MFRL policies with accessibility constraints ($p>0$) achieve stable service accessibility in the studied zones, whereas the other policies yield a large variance among 10 evaluation runs. When demand shocks are introduced, our proposed policies still maintain high accessibility. However, it should be noted that a zone is considered ``accessible'' if at least one available vehicle is located inside the zone. Hence, the result in \Cref{fig:performance_shock_demand_accessibility} could be misleading as it does not reflect the supply-demand relationship, which is better illustrated in \Cref{fig:performance_shock_demand_service_rate}. It can be observed that under the historical demand, all requests are served regardless of the vehicle rebalancing policy. The service rate, however, drops significantly when demand shocks appear. Specifically, in the mild shock with 1,300 requests, no rebalancing and LP approaches leave 60-70\% requestes unserved. In contrast, the proposed policies manage to keep a service rate between 70\% and 85\%. In the extreme demand shock with 5,200 requests, the service rate further plummets, and even the proposed policies cannot serve half of the requests. In both scenarios, \mfrlp{0.85} performs the best thanks to the more uniform distribution of vehicles induced by the accessibility constraint. \looseness=-1

\Cref{fig:performance_shock_demand_pickup_distance} plots the pickup distance in different demand scenarios. As expected, the proposed policies outperform benchmark policies in both demand shocks with shorter pickup distances, though the advantage almost diminishes in the extreme demand shock. Besides, the increase in pickup distance from historical demand to the mild demand shock is more significant than the drop in service rate. This result implies that the accessibility constraint works effectively in guaranteeing service accessibility but can hardly support the same level of service (which is often measured by the rider waiting time and, equivalently, the pickup distance). 

In \Cref{fig:performance-shock_demand}, we also plot the results of a profit-driven MFRL policy, i.e., \mfrlp{0}, which is trained to maximize profit by closely matching the historical demand pattern. As a result, very few vehicles are present in the zones of demand shock, and thus \mfrlp{0} yields the worst performances among all tested policies. 
This finding demonstrates the importance of the accessibility constraint beyond just considering service accessibility and equity: it also makes the service more robust against unpredictable demand shifts and surges.

\paragraph{Robustness to initialization of vehicles}
As elaborated in \Cref{sec:learning-protocol-setup}, we train policies with the initial vehicle distribution $\mu_0$ in line with the demand distribution $\odelta_0$ when the accessibility constraint is imposed ($p>0$).
Hence, it is beneficial to check the robustness of their performances under various initial vehicle distributions constructed as per \Cref{sec:vehicle_initial_movement}. \looseness=-1

\begin{figure*}[h!]
\centering
\begin{subfigure}[t]{0.49\textwidth}
    \centering
    \includegraphics[width=0.99\textwidth]{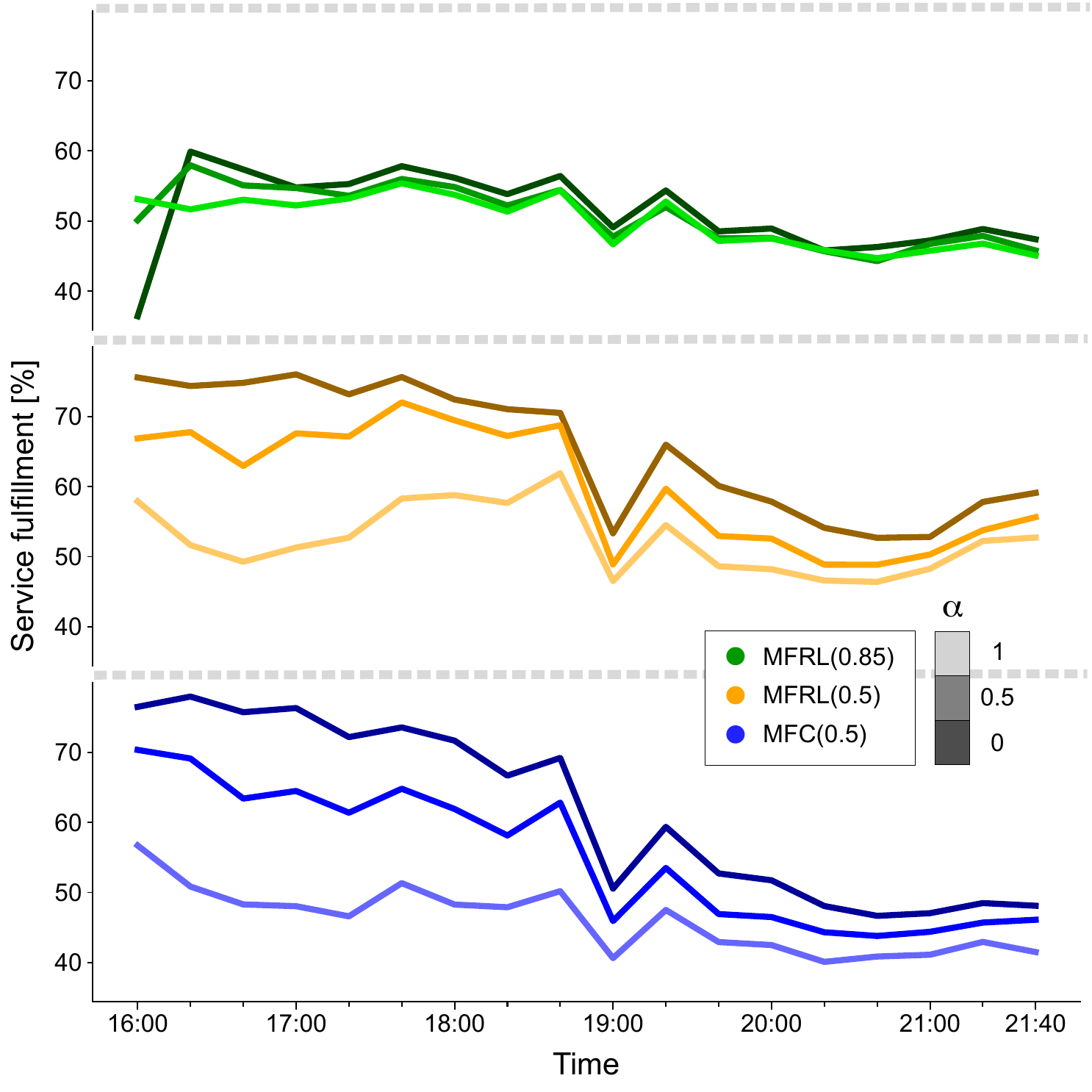}
    \captionsetup{justification=centering}
    \caption{vs. service fulfillment}
    \label{fig:robust-init-fulfill}
\end{subfigure}
\hfill
\begin{subfigure}[t]{0.49\textwidth}
    \centering
    \includegraphics[width=0.99\textwidth]{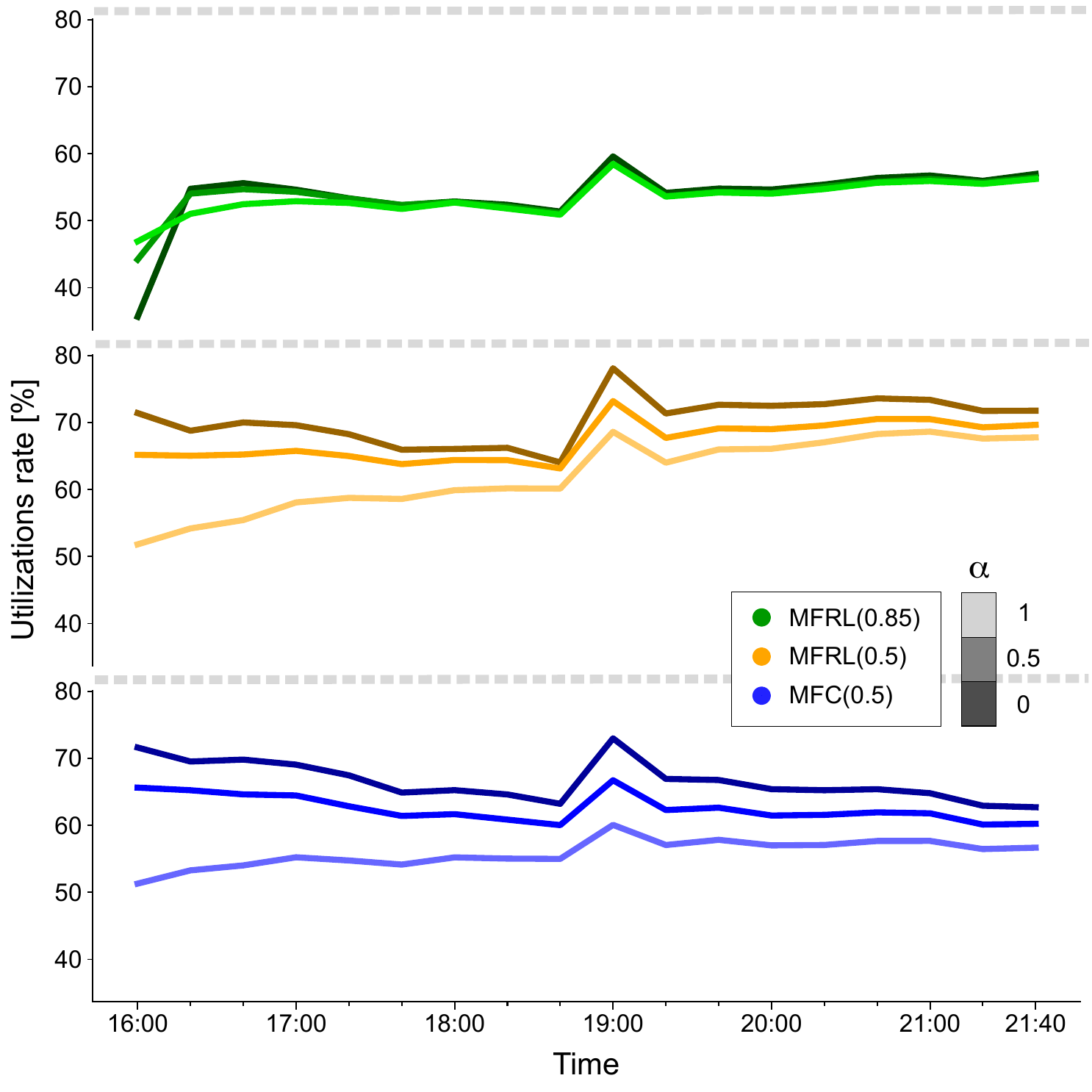}
    \captionsetup{justification=centering}
    \caption{vs. utilization rate}
    \label{fig:robust-init-matched}
\end{subfigure}

\vspace{0.5cm} 

\begin{subfigure}[t]{0.49\textwidth}
    \centering
    \includegraphics[width=0.99\textwidth]{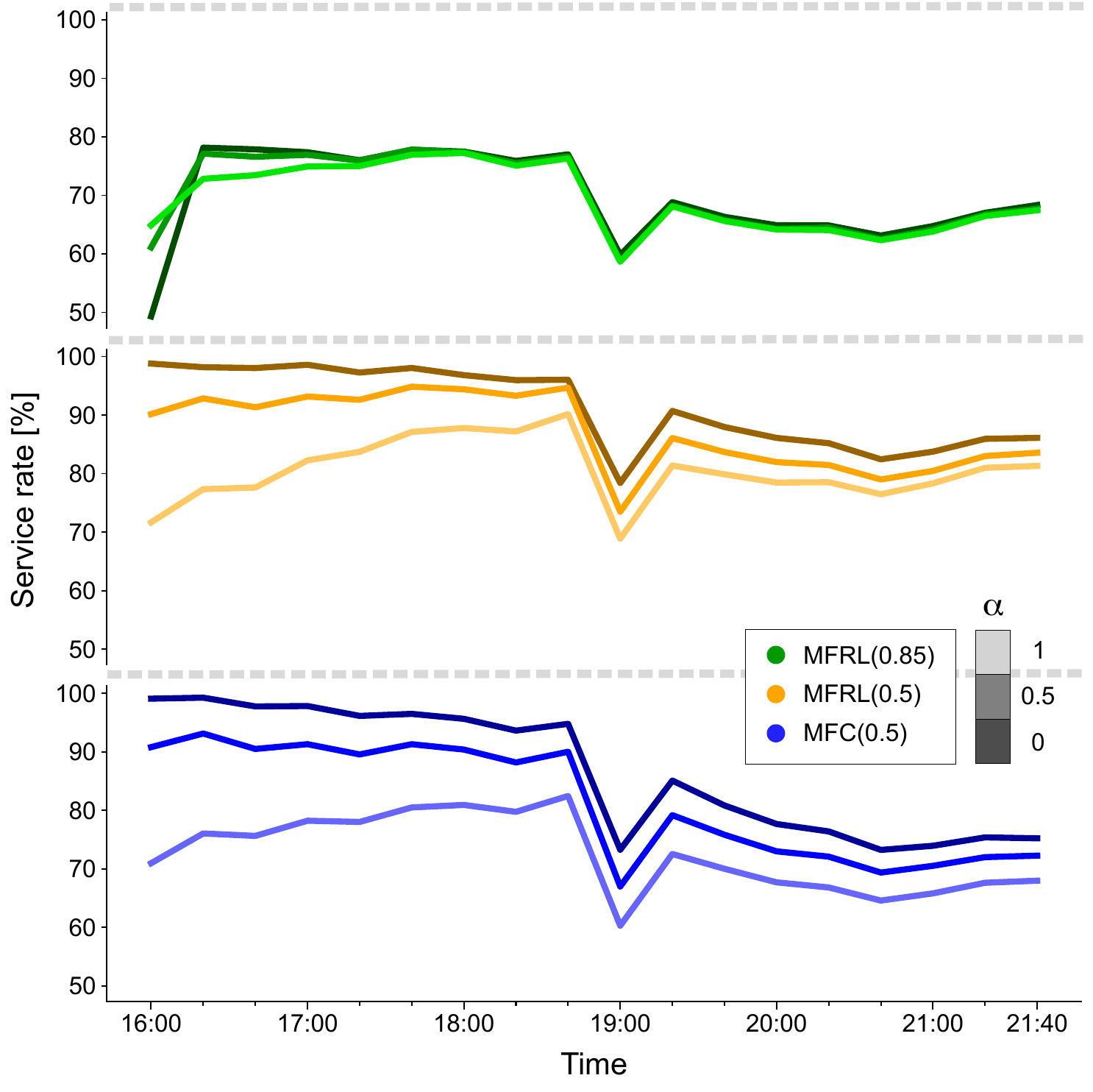}
    \captionsetup{justification=centering}
    \caption{vs. service rate}
    \label{fig:robust-init-satisfied}
\end{subfigure}
\hfill
\begin{subfigure}[t]{0.49\textwidth}
    \centering
    \includegraphics[width=0.99\textwidth]{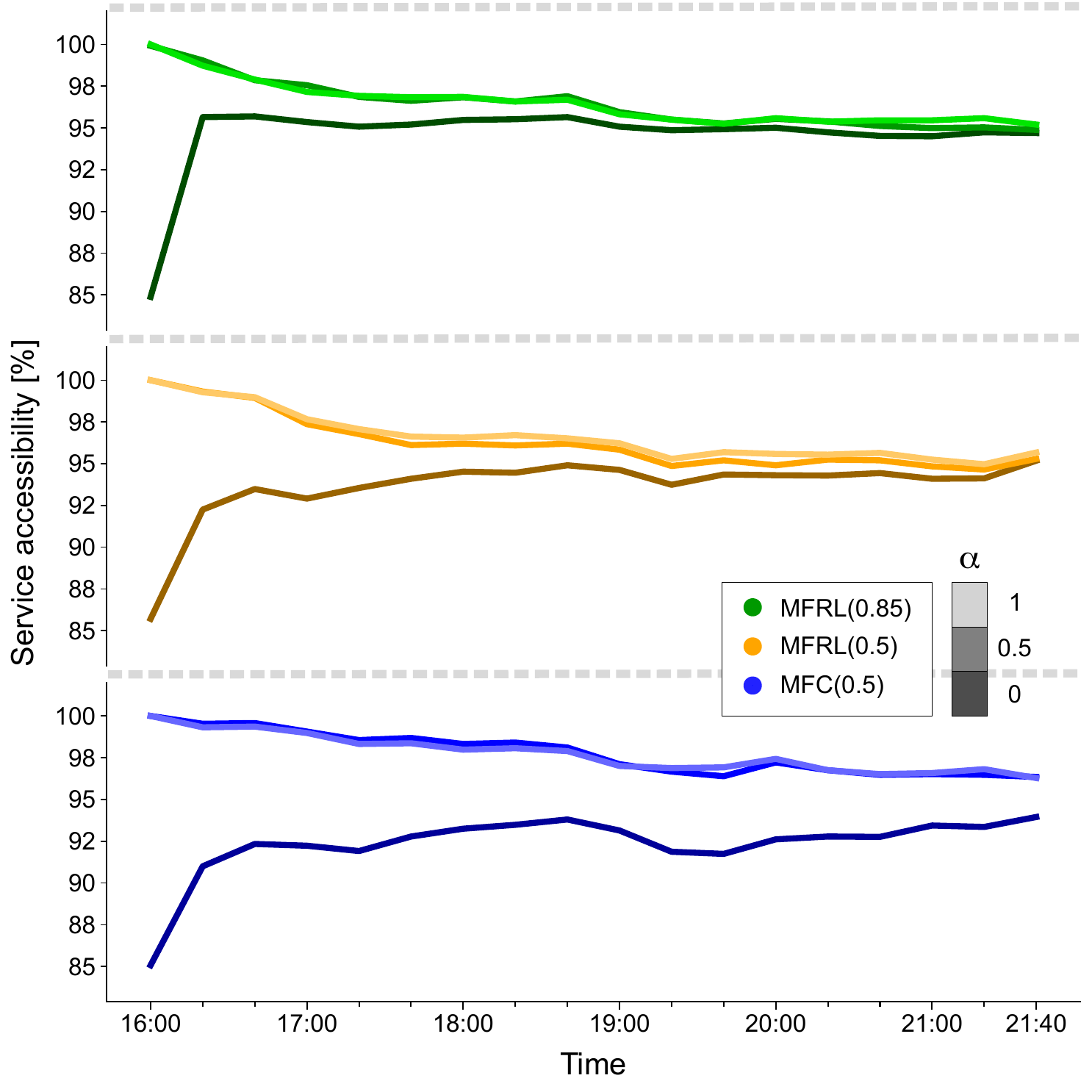}
    \captionsetup{justification=centering}
    \caption{vs. service accessibility}
    \label{fig:robust-init-access}
\end{subfigure}

\caption{Performance metrics over evaluation time under different policies and vehicle initializations.}
\label{fig:robustness_initialization}
\end{figure*}

In \Cref{fig:robustness_initialization}, we analyze how the initial inference-time vehicle distribution $\mu_0$ affects the temporal evolution of performance metrics. Specifically, we plot the performance metrics against time in three vehicle initialization scenarios ($\alpha = 0,0.5,1$). Recall that all tested policies are trained on the vehicle initialization with $\alpha=0$ in \Cref{eq:vehicles_initialization}. Hence, a policy is considered more robust if it yields a smaller gap between the curve of $\alpha=0$ and the others, or the gap quickly converges to zero as time proceeds. \looseness=-1

As can be seen in \Cref{fig:robustness_initialization}, MFRL generally achieves higher robustness than MFC, although the advantage varies across performance metrics. Specifically, in \Cref{fig:robust-init-fulfill}, the service fulfillment drops by almost 20\% for \mfcp{0.5} when vehicles are initialized uniformly ($\alpha=1$), whereas it hardly changes for \mfrlp{0.85} in the same scenario. A similar result is found in Figures~\ref{fig:robust-init-matched} and \ref{fig:robust-init-satisfied} when comparing the utilization and service rates. \mfrlp{0.5} displays similar trends to \mfcp{0.5}, but shows a higher robustness with smaller gaps. 
Nevertheless, all policies manage to converge towards stable system outcomes over time, regardless of the initial vehicle distribution. 

Another interesting observation is that the service accessibility is rather sensitive to vehicle initialization, as shown in \Cref{fig:robust-init-access}. When half of the vehicles are uniformly distributed ($\alpha=0.5$), the resulting service accessibility already reaches the same level as in the scenario of purely uniform vehicle initialization ($\alpha=1$). This result is explained by the zone being considered accessible by having a single available vehicle. In contrast, the shift is more linear and gradual for other performance metrics. \looseness=-1

In sum, our analysis shows that the most conservative \mfrlp{0.85} policy is also more resilient to vehicle initialization. Compared to MFRL, the MFC policy is prone to degraded performance, possibly because it is trained on a fixed approximation of the matching process. In contrast, MFRL learns a more grounded matching process and thus can produce more robust vehicle rebalancing strategies.

\FloatBarrier

\section{Conclusions} \label{sec:conclusion}

This study proposes large-scale ride-sourcing vehicle rebalancing approaches based on continuous-state mean-field control (MFC) and mean-field reinforcement learning (MFRL). The mean-field approximation eliminates the curse of dimensionality in the number of vehicles and enables real-time and adoptive vehicle rebalancing at the metropolitan scale. We further introduce an accessibility constraint into both models and ultimately strike a balance between efficiency and equity. 

Experiment results show that well-trained MFC and MFRL outperform benchmark policies in both efficiency and effectiveness. Particularly, they generate rebalancing actions for a fleet of 18,000 vehicles within one second, whereas the LP-based approaches take more than 10 minutes. MFC and MFRL also achieve higher service fulfillment and vehicle utilization with fewer rebalancing trips. In general, MFRL performs better than MFC with up to 5\% margin in most performance metrics, though it is also more computationally expensive. 
Importantly, both MFC and MFRL achieve satisfactory service accessibility without a significant drop in service efficiency. Specifically, the two policies are able to explore the efficiency-equity Pareto front more thoroughly, yielding solutions that dominate conventional benchmarks across critical service metrics, such as fleet utilization rate, the number of fulfilled ride requests, and pickup distance, meanwhile ensuring equitable service access. 
Furthermore, the proposed approaches show strong robustness towards unpredictable demand shifts and extreme demand shocks with 70–85\% service rates, meanwhile benchmark policies fail to serve more than 40\% of the requests.

Although the rebalancing approaches proposed in this paper show strong potential for real-world deployment, several important theoretical, engineering, and societal challenges remain. For instance, we have assumed that the ride-sourcing platform can control the vehicle rebalancing with full compliance. This assumption does not hold in the current practice with human drivers, and bridging this gap would require sophisticated modeling of driver behaviors. However, this limitation is expected to diminish as autonomous fleets become more prevalent. Another simplification introduced in this paper regards the time discretization. It has been assumed that all trips are completed within a single interval, while allowing multi-step flows~\citep [e.g.,][]{zhang2023ride} should not largely change the modeling framework and computational complexity. 
Since this paper mainly focuses on addressing the scalability issue with respect to fleet size, the service region and time horizon are restricted in the experiments. Future research shall test the proposed algorithms over larger areas and longer time periods. 
Moreover, the current accessibility constraint is imposed at the system level, while finer-grained and local constraints shall be introduced in future research to derive more context-aware rebalancing policies.
Last but not least, it is worthwhile to extend the current single-platform framework to capture competition among multiple fleets.

\section*{Acknowledgments}
We want to thank Ilija Bogunovi\'c for insightful discussions during the initial phase of the project. We would also like to thank the Shenzhen Urban Transport Planning Center for collecting and sharing the data used in this work.
Zhiyuan Hu acknowledges support from the Swiss National Science Foundation under the research grant PZ00P2\_216211. 
This publication was made possible by an ETH AI Center doctoral fellowship to Barna P\'asztor.
Andreas Krause acknowledges that this research was supported by the European Research Council (ERC) under the European Union’s Horizon 2020 research and innovation program grant agreement no. 815943 and the Swiss National Science Foundation under NCCR Automation, grant agreement 51NF40180545.
Matej Jusup and Francesco Corman acknowledge support from the Swiss National Science Foundation under the research project DADA/181210. 
Finally, Kenan Zhang acknowledges support from the Swiss National Science Foundation under the research grant 200021\_219232. 
\looseness=-1

\clearpage
\newpage
\printbibliography

\newpage

\appendix

\section{Lifted \cmfmdp} \label{apx:lifted-c-mf-mdp}
In general, the expectation in \Cref{eq:mfc} does not have an analytical solution, while sampling-based approximations would be impractical and computationally expensive.
To make the optimization in \Cref{eq:mfc} feasible in practice, we utilize a lifted MF-MDP reformulation \citep[e.g.,][]{carmona2019model, Gu2019DynamicMFCs, Gu2021Mean-FieldAnalysis, Motte2019Mean-fieldControls, pasztor2021efficient} that converts \Cref{eq:mfc} into an iterative procedure. \citet{jusup2024safe} introduce lifted \cmfmdp and associated log-barrier variant, which we use to optimize \Cref{eq:log-barrier} iteratively. In general, the first step is to define the lifted reward as an integral of the reward over the state space with respect to the mean-field probability measure:
\begin{equation*}
    \oR(\mut, \deltat, \pit) {}={} \int_{s\in\s} r(\mut, \deltat, \pit(\st, \mut, \deltat))\mut(\ds).
\end{equation*}
Approximating the above integral is often much easier, especially in low-dimensional state spaces, than computing the original expectation.

We now reformulate \Cref{eq:log-barrier} as:
\begin{subequations}\label{eq:lifted-log-barrier}
\begin{align}
    \bpis \eq\argmax_{\bpi \in \bPi} {}\; & \sum_{t=0}^{T-1} \oR(\mut, \deltat, \pit) + \lambda\log(\hw(\mutn, \pitn) - C) \\
    \text{s.t.} \quad \mutn &\eq\Umpf. \label{eq:lifted-log-barrier-mf-transition}
\end{align}
\end{subequations}
Intuitively, \cmfmdp is lifted from the representative vehicle perspective to the population/system-level perspective. Consequently, the noise is integrated out, making $U(\cdot)$ deterministic.
However, the reformulation comes at the expense of more complex state and action spaces, $\s_{\text{new}}=\pps$ and $\A_{\text{new}}=\{\pi: \s \times \pps \times \mms \to \A\}$. In the implementation, we discretize $\s_{\text{new}}$ in a grid and define the ``space of policies'' $\A_{\text{new}}$ over the grid midpoints. \looseness=-1

\section{Optimal Transport as the Minimum Cost Flow} \label{apx:discrete-ot}
\begin{figure*}[htb!]
\centering
\begin{subfigure}[t]{\textwidth}
    \centering
    \includegraphics[width=\textwidth, trim=1.68cm 10cm 1.68cm 10cm, clip]{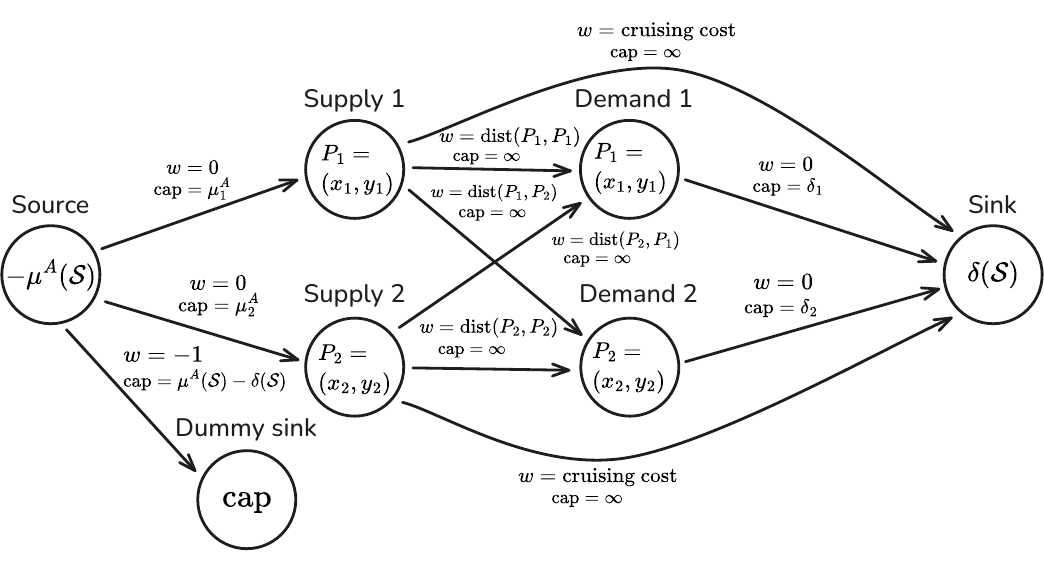}
    \captionsetup{justification=justified}
    \caption{The optimal transport graph when supply exceeds demand.}
    \label{fig:ot-higher-supply}
\end{subfigure}
\par\vspace{0.3cm}
\begin{subfigure}[t]{\textwidth}
    \centering
    \includegraphics[width=\textwidth, trim=1.68cm 10cm 1.68cm 10cm, clip]{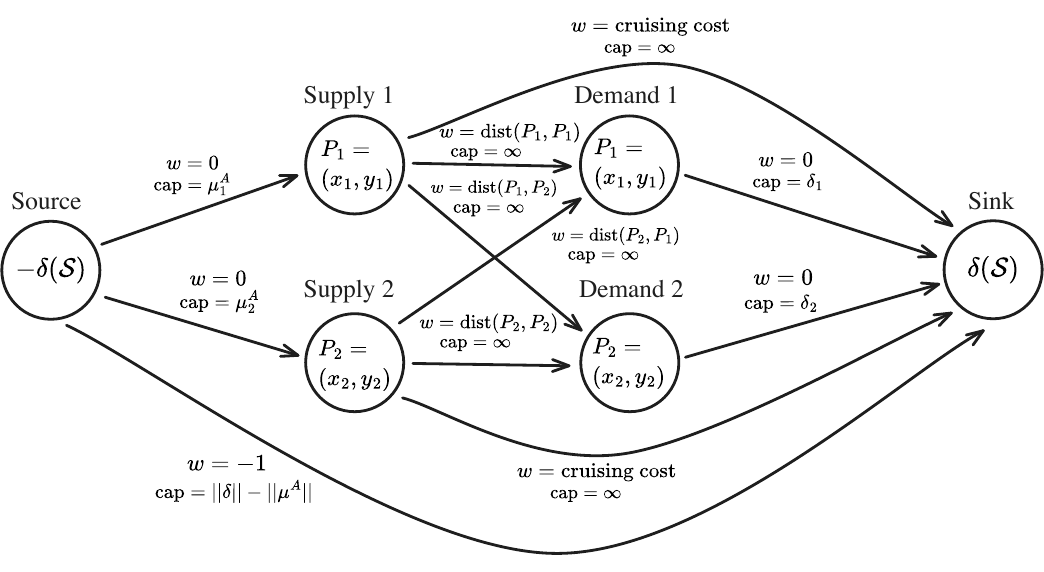}
    \captionsetup{justification=justified}
    \caption{The optimal transport graph when demand exceeds supply.}
    \label{fig:ot-higher-demand}
\end{subfigure}
\caption{The optimal transport as the minimum cost flow, which, as a sub-problem, solves bipartite matching between supply and demand measures. We distinguish two cases: (a) supply exceeds (or equals) demand, (b) demand exceeds supply. \looseness=-1}
\label{fig:ot-networks}
\end{figure*}

In \Cref{eq:matching_process_approx}, we introduced the optimal transport as an approximation of the matching process. Working directly with continuous measures is often hard, so in \Cref{sec:state-space-representation}, we opted for a two-dimensional grid representation over the city map. Intuitively, the grid splits the city into $N$ zones with zone center coordinates $P_i=(x_i,y_i), i=1,2,\ldots,N$. Each zone has associated supply and demand measures, i.e., measures of available vehicles $\mua=(\mua_1,\mua_2, \ldots, \mua_N)$ and rider requests $\delta=(\delta_1,\delta_2, \ldots, \delta_N)$. This representation gives us the possibility to approximate the continuous optimal transport from \Cref{eq:matching_process_approx} as the minimum cost flow problem~\citep{ahuja1993network,ford2015flows} with nodes representing zone-level supply and demand ``masses.'' Intuitively, the goal is to transfer all the supply measure $\mua(\s)=\sum_{i=1}^N\mua_i$ from the source node to all the demand measure $\delta(\s)=\sum_{i=1}^N\delta_i$ at the sink node across the state space $\s$. The ``inner'' nodes of the network are constructed as a bipartite matching between zone-level supply ($\mua_i$) and demand ($\delta_i$) measures. The global matching would be unreasonable (long waiting times) and computationally expensive ($\bigO(N^2)$ for a fully-connected graph), so we connect zones only to their direct neighbors (and themselves). The cost (i.e., edge weight) of transferring mass from a supply zone to a demand zone is the Euclidean distance between the zone centers. We also include the cruising edges with manually set cruising costs. We set the cruising cost to 2, which is around forty times higher than the maximum Euclidean distance of neighboring zones -- the diagonal neighbors have a distance of around 0.05 because the map is represented as the square $[0,1]^2$. 
The final hurdle comes from the fact that the supply measure $\mua(\s)$ and the demand measure $\delta(\s)$ might not be equal. In the case of $\delta(\s) \leq \mua(\s)$, we forward excessive supply to a dummy sink as depicted in \Cref{fig:ot-higher-supply}. In the case of $\mua(\s) < \delta(\s)$, we set the source capacity to the negative demand measure and forward the excessive demand directly to the sink to preserve the flow balance, as depicted in \Cref{fig:ot-higher-demand}. Note that in \Cref{fig:ot-networks} we showcase only a simple network with two neighboring zones, but it is straightforward to scale it up to $N>2$ following the above instructions.

\clearpage
\section{Benchmarks} \label{apx:benchmarks}
We use two linear programs as benchmarks.

\subsection{Static Rebalancer} \label{apx:static-rebalancer}
\subsubsection{Motivation} 
The static rebalancer aims to keep the supply at each zone constant by rebalancing vehicles to compensate for the difference between incoming and outgoing requests at each zone.

\subsubsection{Formulation} We use the linear programming formulation of the static rebalancer introduced in~\citet{pavone2012robotic}: \looseness=-1

\begin{align}
    \min_{\alpha_{ij}} \, & \sum_{i,j} T_{ij}\alpha_{ij} \\
    \text{s.t.} \; & \sum_{i \neq j} (\alpha_{ij} - \alpha_{ji}) = - \lambda_{i} + \sum_{i \neq j}\lambda_{j} p_{ji}, \, \forall i \in \mathcal{N} \label{eq:static-rebalancer-c1} \\
    & \alpha_{ij} \geq 0, \, \forall i,j \in \mathcal{N}. \label{eq:static-rebalancer-c2}
\end{align}
In \Cref{tab:static-rebalancer}, we list the model variables and their descriptions.
\Cref{eq:static-rebalancer-c1} in the formulation ensures that the total outgoing rebalancing rates should equal the total incoming request rates for each node, and \Cref{eq:static-rebalancer-c2} ensures nonnegativity. \looseness=-1

\begin{table*}[hbt!]
\centering
\caption{Static rebalancer variables}
\label{tab:static-rebalancer}
    \begin{tabular}{c|l}
        \toprule
        Variable & \multicolumn{1}{c}{Description} \\
        \midrule
        $\mathcal{N}$ & A set of nodes. E.g., zones in the city. \\
        $\alpha_{ij}$ & Rebalancing rate from node $i$ to node $j$ \\
        $T_{ij}$ & Weight to model the effort to rebalance from node $i$ to node $j$ \\
        $\lambda_{i}$ & Total request rates from node $i$ \\
        $p_{ij}$ & The probability of a request ending in node $j$ given it starts at node $i$ \\
        \bottomrule
    \end{tabular}
\end{table*}

\subsubsection{Technical Details} \label{apx:static-rebalancer-technical-details}
In the episodic setting, the static rebalancer is solved iteratively for every time step. We set $T_{ij}$ uniform for every $i,j$, assuming a uniform rebalance effort for every OD pair.

\noi\emph{Inputs}: Total count of requests from every node and the corresponding ending probabilities. The parameter $\lambda_{i}$ is estimated as the count divided by the length of the time step, and $p_{ij}$ directly equals the ending probabilities.

\noi\emph{Outputs}: The rebalancing rate $\alpha_{ij}$, indicating how many vehicles to reposition per time unit. These rates are further multiplied by the time length to return the final count of rebalancing vehicles for one time step.

\subsubsection{Limitations} 
Some of the limitations of this formulation are:

\begin{enumerate}
    \item It assumes constant request rates, which may be an oversimplification for real-world requests.
    \item It assumes a uniform rebalancing effort, which omits the difference between different OD pairs.
    \item For the episodic setting, $\lambda_{i}$ should not be directly used to compute the incoming rates at the destinations since a trip starting in a time step may not end in the same one. A more realistic way could be having two versions of $\lambda_{i}$ based on start and end time, but it will harm the feasibility of the program.
    \item It does not consider the available supply at each node.  
\end{enumerate}

\subsection{Dynamic Rebalancer} \label{apx:dynamic-rebalancer}

\subsubsection{Motivation}
In the real world, the request distribution may change with respect to time, and it may not be optimal to keep the supply distribution unchanged but adjust it according to the real-time request condition. Instead of assuming a constant request rate and keeping the supply exactly constant throughout time steps, the dynamic rebalancer also takes in the supply information and aims to keep the supply above a manually defined threshold.

\subsubsection{Formulation}
We use the linear programming formulation of the dynamic rebalancer introduced in~\citet{pavone2012robotic}: \looseness=-1
\begin{align}
    \min_{\text{num}_{ij}} \, & \sum_{i,j} T_{ij} \text{num}_{ij} \\
    \text{s.t.} \; & v_i^{ex} + \sum_{i \neq j} (\text{num}_{ij} - \text{num}_{ji}) \geq v_i^d, \, \forall i \in \mathcal{N} \label{eq:dynamic-rebalancer-c1} \\
    & \alpha_{ij} \in \mathbb{N}, \, \forall i,j \in \mathcal{N}. \label{eq:dynamic-rebalancer-c2}
\end{align}
In \Cref{tab:dynamic-rebalancer}, we list the model variables and their descriptions.
Compared to the static rebalancer, the dynamic rebalancer directly takes in the count of requests and vehicles. \Cref{eq:dynamic-rebalancer-c1} ensures that the new supply, after rebalancing, should not fall below the desired threshold.

Although \Cref{eq:dynamic-rebalancer-c2} requires the decisions to be integers, there are theoretical guarantees that it can be relaxed to real numbers. Thus, we can solve the program as a linear program, and the optimal decisions will always be integers. The parameters $v_i^{ex}$ and $v_i^d$ can be flexibly set with the supply and request situation, making it easier to adjust to different desires. \looseness=-1

\begin{table*}[htb!]
\centering
\caption{Dynamic rebalancer variables}
\label{tab:dynamic-rebalancer}
    \begin{tabular}{c|l}
        \toprule
        Variable & \multicolumn{1}{c}{Description} \\
        \midrule
        $\mathcal{N}$ & A set of nodes. E.g., zones in the city. \\
        $\text{num}_{ij}$ & Number of vehicles to reposition from node $i$ to node $j$ \\
        $T_{ij}$ & Weight to model the effort to reposition from node $i$ to node $j$ \\
        $v_{i}^{ex}$ & Number of excessive vehicles at node $i$ which can be used for rebalancing \\
        $v_{i}^{d}$ & Desired number of vehicles at node $i$ after rebalancing \\
        \bottomrule
    \end{tabular}
\end{table*}

\subsubsection{Technical Details} \label{apx:dynamic-rebalancer-technical-details}
During simulation, the excessive vehicle $v_i^{ex}$ is computed as the initial supply at node $i$ minus the count of requests leaving node $i$; if the result is negative, then we set $v_i^{ex}$ to 0. For $v_i^{d}$, it is set as 80\% of the initial supply minus the count of vehicles entering node $i$. By this setting, we ensure that the updated initial supply at the next time step will not fall below 80\% of the initial supply at the current time step.

\noi\emph{Inputs}: $v_i^{ex}$ and $v_i^{d}$ computed by the initial supply and request counts.

\noi\emph{Outputs}: $\text{num}_{ij}$, which directly indicates how many vehicles to reposition for every OD pair.

\FloatBarrier
\subsubsection{Limitations}
Some of the limitations of this formulation are:

\begin{enumerate}
    \item It still assumes a uniform rebalancing effort for every OD pair.
    \item It does not ensure that the rebalancing decisions will not exceed the capacity $v_i^{ex}$ in the first place. An improvement can be adding the constraint $v_i^{ex} \geq \sum_{j} \text{num}_{ij}, \forall i \in \mathcal{N}$, but then the program can not be solved as an LP, which makes it significantly more difficult.
    \item The flexibility in choosing the parameters also makes it difficult to determine the optimal computation for the parameters. \looseness=-1
\end{enumerate}

\section{Impact of the Log-Barrier Hyperparameter $\lambda$} \label{apx:impact-of-lambda}

In \Cref{sec:log-barrier-method}, we used the log-barrier method to relax the original formulation in \Cref{eq:mfc}. This replaces the hard constraint \Cref{eq:mfc-safety} with a soft log-barrier term, weighted by $\lambda$, in the objective function. The advantage of this reformulation is that it enables the use of standard RL optimization techniques. The drawback is that the resulting policy may differ from the original constrained solution because: (i) the logarithmic penalty strongly discourages solutions near the boundary, and (ii) performance depends on the choice of the hyperparameter $\lambda$.
This section investigates how $\lambda$ influences training, specifically the convergence of the learning protocol (\Cref{alg:learning_protocol}). To better understand this behavior, we analyze the trade-off in the log-barrier objective (\Cref{eq:log-barrier-objective}) between the reward $r(\cdot)$ and the logarithm of the accessibility constraint slack $\log(\hw(\cdot) - C)$. Results in \Cref{tab:impact-of-lambda} confirm the intuition: for small $\lambda$, optimization favors increasing $r(\cdot)$ since the log-barrier penalty is negligible, whereas for large $\lambda$, optimization prioritizes increasing the slack to avoid large penalties.
To our knowledge, there is no general method for selecting an optimal $\lambda$, but understanding problem structure can help avoid costly hyperparameter tuning. In our case, $r(\cdot)$ lies in the $[0,1]$ interval, while the log-barrier term ranges over $(-\infty, x]$, where $x$ depends on the maximum entropy $\hw_{\max}$ and the fraction $p$ of $\hw_{\max}$. For example, when $p=0.05, 0.5, 0.85$, we obtain $x=1.77, 1.12, -0.08$, respectively. For simplicity, we fix $\lambda=1$ in all experiments, which places the log-barrier term on a scale comparable to the reward. Empirical results confirm that this choice yields high-performing (see \Cref{sec:results}), albeit conservative, policies that satisfy the accessibility constraint (\Cref{apx:constraint-satisfaction}). However, for business-critical applications, $\lambda$ can be further tuned.

\begin{table*}[htb!]
\centering
\caption{Impact of log-barrier hyperparameter $\lambda$ on reward $r(\cdot)$ and logarithm of the accessibility constraint slack $\log(\hw(\cdot)-C)$ during training, averaged over $T=18$ steps. Small $\lambda$ favors reward, large $\lambda$ favors the log-barrier term, and $\lambda=1$ keeps the two on a comparable scale.}
\label{tab:impact-of-lambda}
\small
\begin{tabular}{l|l|*{5}{c}}
\toprule
Model Name & \diagbox{Term}{$\lambda$} & $0.01$  & $0.1$  & 1 & $10$  & $100$ \\
\midrule
\multirow{2}{*}{\mfrlp{0.5}} & $r(\cdot)$ & 0.67 & 0.66 & 0.61 & 0.46 & 0.24  \\
                             & $\log(\hw(\cdot) - C)$ & 0.83 & 0.86 & 0.93 & 0.98 & 0.98 \\
\midrule
\multirow{2}{*}{\mfrlp{0.85}} & $r(\cdot)$ & 0.67 & 0.61 & 0.53 & 0.26 & 0.23 \\
                            & $\log(\hw(\cdot) - C)$ & -1.45 & -0.91 & -0.72 & -0.62 & -0.68 \\
\bottomrule
\end{tabular}
\end{table*}

\section{Accessibility Constraint Satisfaction} \label{apx:constraint-satisfaction}

\begin{figure*}[htb!]
\centering
\includegraphics[width=0.5\textwidth]{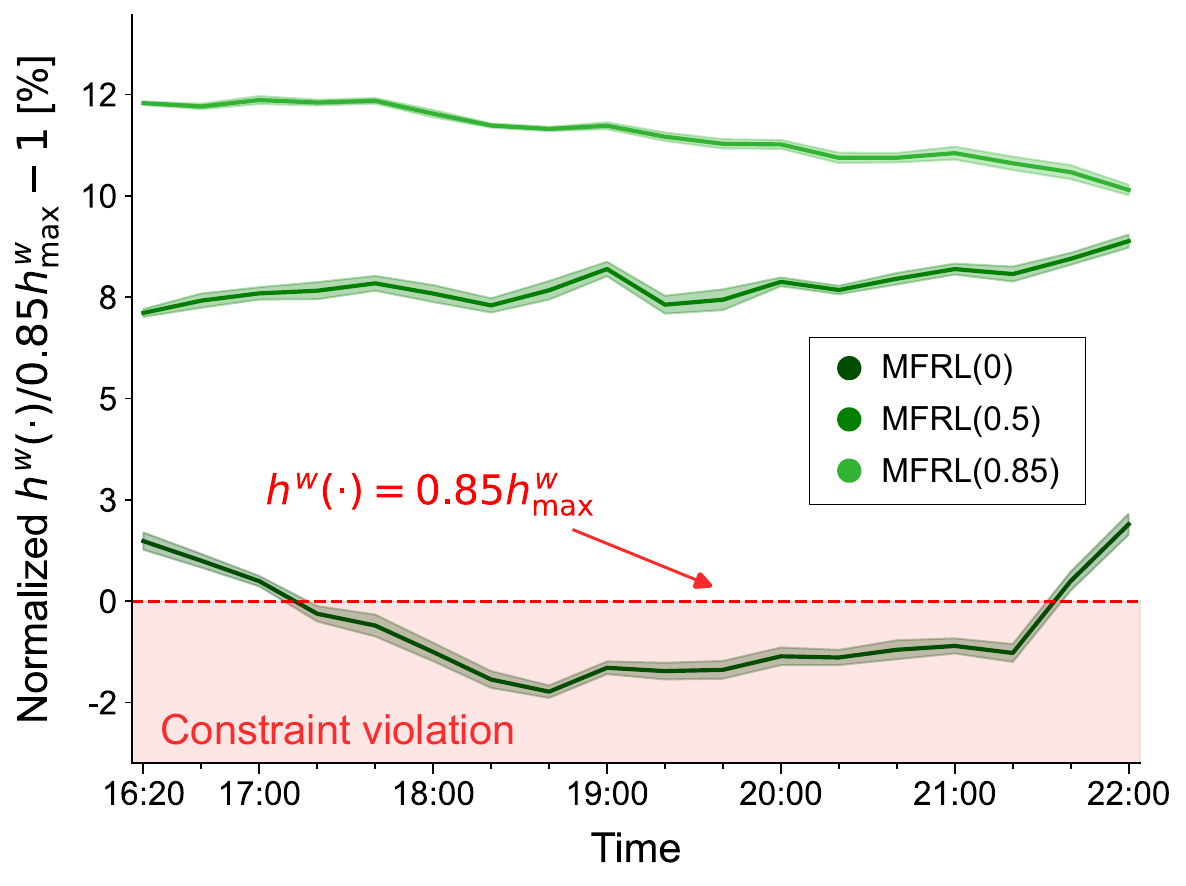} 
\caption{Accessibility constraint satisfaction under MFRL policies. The plot shows normalized slacks $\nicefrac{\hw(\cdot)}{C} - 1$ (mean: solid green, $\pm$1 std: shaded) across 10 runs, relative to the binding constraint $\hw(\cdot)=C$ (dashed red) and violation region $\hw(\cdot)<C$ (shaded red).} \label{fig:constraint_satisfaction}
\end{figure*}

The goal of this section is to verify that fleets following the learned policy $\bpi_N^*$ (\Cref{alg:learning_protocol}) satisfy the accessibility constraint in \Cref{eq:weighted-accessibility-constraint} during evaluation.
In \Cref{fig:constraint_satisfaction}, we report the normalized slack $\nicefrac{\hw(\cdot)}{C} - 1$ between the accessibility function $\hw(\cdot)$ (\Cref{eq:weighted-differential-entropy}) and the lower bound $C=0.85h_{\max}^w$, averaged over 10 random seeds. The slack quantifies conservativeness: larger values indicate safer policies.
As expected, \mfrlp{0.85} always satisfies $\hw(\cdot) \geq C$, while the unconstrained model \mfrlp{0} violates the requirement almost immediately. To highlight conservativeness, we also train \mfrlp{0.5} and evaluate it against the stricter threshold $p=0.85$, i.e., $C=0.85h_{\max}^w$. Results confirm that accessibility is consistently maintained, with smaller safety margins than \mfrlp{0.85}.

\section{Spatiotemporal Demand Variability} \label{apx:demand-variability}
\begin{figure*}[htb!]
\centering
\includegraphics[width=0.5\textwidth]{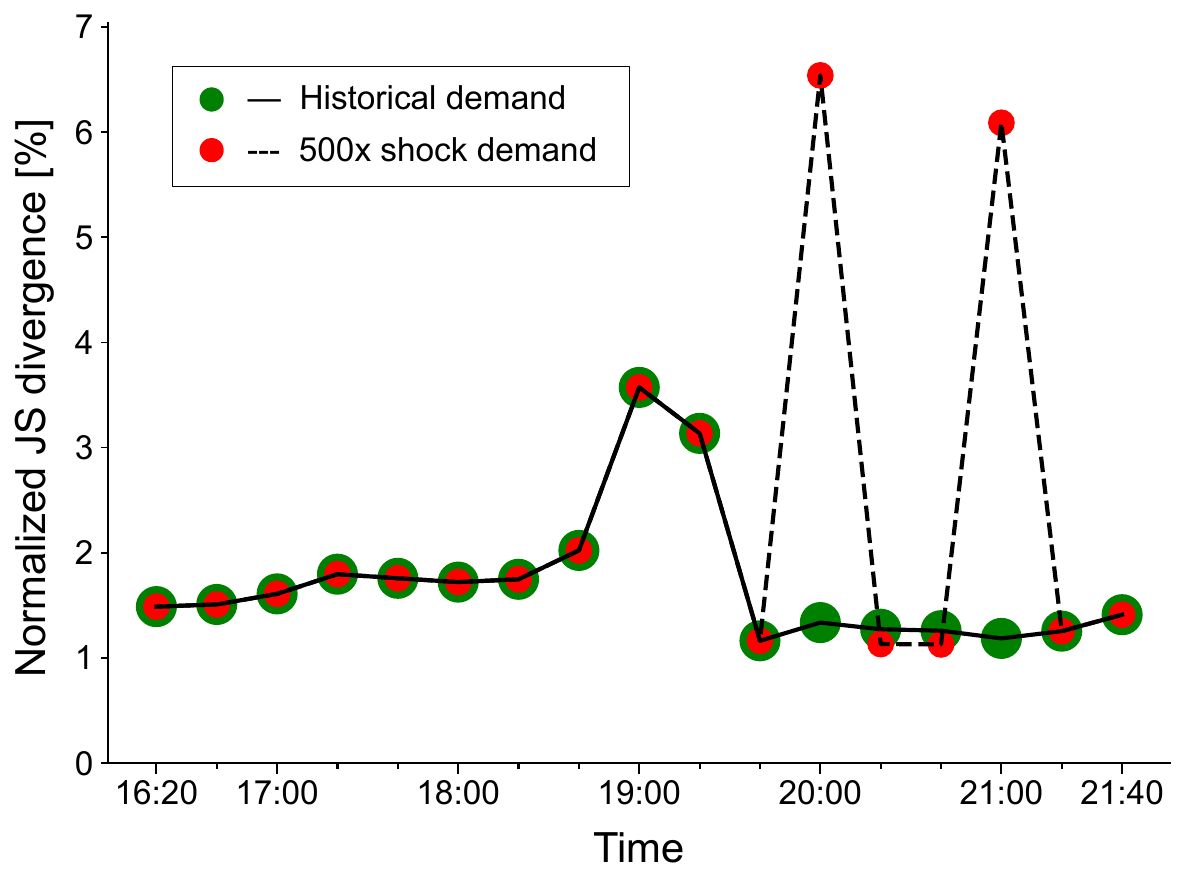} 
\caption{Historical and hypothetical demand variability (500x shock scenario) over the study period 16:00-22:00.} \label{fig:demand_variability}
\end{figure*}

While there is no conceptual obstacle to applying our model over the full 0–24 hour interval by increasing the number of time steps, performing an exhaustive analysis---training and evaluating multiple models across numerous random seeds---exceeds our current compute budget. Furthermore, the slow runtime of LP benchmarks prevents extending them to the full interval, leaving only the trivial ``No Rebalancing'' baseline for comparison.
Consequently, we limit our service period to the evening peak hours, 16:00–22:00, divided into 20-minute intervals. First, peak hours are of primary interest to service providers. Second, this period exhibits notable---though still limited---spatiotemporal demand variability, as shown in \Cref{fig:demand_variability}. Variability is measured using the Jensen–Shannon (JS) divergence, which captures changes in demand both across zones and between consecutive time periods. Since JS divergence is bounded in $[0, \log 2]$, we report a normalized version, $\text{JS} / \log 2$, so that values lie in $[0,1]$ and can be interpreted as percentage changes: 0 indicates identical demand distributions, while 1 corresponds to the maximum possible difference.
Historical variability averages around 1.5\% with a maximum of approximately 4\%. As expected, the hypothetical 500$\times$ shock demand increases variability from \textasciitilde1\% to \textasciitilde7\%, before returning to normal after the one-hour shock period.

\end{document}